%% file: main.tex
\documentclass[10pt,twocolumn,letterpaper]{article}

\usepackage[pagenumbers]{wacv}

\input{preamble}
\usepackage[accsupp]{axessibility}  %

\definecolor{wacvblue}{rgb}{0.21,0.49,0.74}
\usepackage[pagebackref,breaklinks,colorlinks,allcolors=wacvblue]{hyperref}

\def\wacvPaperID{3093}
\def\confName{WACV}
\def\confYear{2027}

\title{AnchorUp: Anchoring Feature Upsampling on a Classical Interpolation}

\author{Kotaro Horikawa$^{1}$\quad Keisuke Toida$^{2}$\quad Junya Komori$^{2}$\quad Kazuhiro Hotta$^{1}$\\
$^{1}$Meijo University, Japan\quad $^{2}$Asia Air Survey Co., Ltd., Japan\\
{\tt\footnotesize 263427041@ccmailg.meijo-u.ac.jp, \{ksk.toida, jny.komori\}@ajiko.co.jp, kazuhotta@meijo-u.ac.jp}
}

\begin{document}
\AddToShipoutPicture*{\AtPageUpperLeft{\raisebox{-0.42in}{\makebox[\paperwidth]{\parbox[t]{\textwidth}{\scriptsize\centering \copyright~2027 IEEE. Personal use of this material is permitted. Permission from IEEE must be obtained for all other uses, in any current or future media, including reprinting/republishing this material for advertising or promotional purposes, creating new collective works, for resale or redistribution to servers or lists, or reuse of any copyrighted component of this work in other works.}}}}}
\maketitle

\input{sec/0_abstract}
\input{sec/1_intro}
\input{sec/2_related}

\input{sec/3_method}
\input{sec/4_experiments}
\input{sec/5_analysis}
\input{sec/6_conclusion}

{
    \small
    \bibliographystyle{ieeenat_fullname}
    \bibliography{refs}
}

\setcounter{section}{0}\setcounter{figure}{0}\setcounter{table}{0}\setcounter{equation}{0}
\renewcommand{\thesection}{\Alph{section}}
\renewcommand{\thefigure}{S\arabic{figure}}
\renewcommand{\thetable}{S\arabic{table}}
\renewcommand{\theequation}{S\arabic{equation}}
\renewcommand{\theHsection}{supp.\arabic{section}}
\renewcommand{\theHfigure}{supp.\arabic{figure}}
\renewcommand{\theHtable}{supp.\arabic{table}}
\renewcommand{\theHequation}{supp.\arabic{equation}}
\twocolumn[{\centering\Large\bfseries Supplementary Material\par\vspace{1.5em}}]
\input{sec/7_supp}

\end{document}

%% file: preamble.tex
\usepackage{amsmath}
\usepackage{amssymb}
\usepackage{booktabs}
\usepackage{algorithm}
\usepackage{algpseudocode}
\usepackage{graphicx}
\usepackage{xcolor}
\usepackage{tikz}

\definecolor{deltagreen}{rgb}{0.00,0.50,0.00}
\definecolor{deltared}{rgb}{0.75,0.00,0.00}
\newcommand{\dpos}[1]{\textcolor{deltagreen}{$+$#1}}
\newcommand{\dneg}[1]{\textcolor{deltared}{$-$#1}}
\usetikzlibrary{arrows.meta,positioning,calc}

\newcommand{\Flr}{F_{\mathrm{LR}}}
\newcommand{\Fhr}{F_{\mathrm{HR}}}
\newcommand{\Fhat}{\hat{F}_{\mathrm{HR}}}
\newcommand{\Fanchor}{F_{\mathrm{anchor}}}
\newcommand{\Flearn}{F_{\mathrm{learn}}}
\newcommand{\method}{AnchorUp}

%% file: sec/0_abstract.tex
\begin{abstract}
Vision foundation models produce semantically strong features on a coarse patch
grid, and dense prediction needs them at pixel resolution. Existing learned
upsamplers close that gap by regressing the high-resolution feature map from
the low-resolution one, typically through cross-attention driven by the input
image. We observe that most of what they regress needs no learning: plain
bilinear interpolation already reproduces the smooth regions of the target, and
recent benchmarks even report bilinear and bicubic on par with several learned
upsamplers. The rest is mostly edges and fine details,
which suggests that these learned upsamplers spend much of their capacity
re-computing the smooth content that classical interpolation provides. We
therefore introduce \method{}, which stops predicting the high-resolution
feature map and predicts only its correction from the interpolation. It keeps a parameter-free classical upsampler as an
anchor, obtains a second prediction from a small guidance network, and blends
the two through a single learnable scalar gate.
Because the anchor supplies the smooth content, the guidance network
only has to correct what remains. An analysis of the learned correction
confirms that it concentrates on edges and fine details, and so the
network can be small: \method{} uses 167{,}425 parameters, a quarter of the budget of the
strongest baseline, NAF, and runs at 64\% of its latency. Trained once on
DINOv3 with only the first of NAF's two training stages, and evaluated on that
backbone plus six held-out ones, it is already on par with NAF trained
through both stages on Pascal VOC and Cityscapes. Our method surpasses it on DAVIS-2017 video object
segmentation;
the gain holds on every
backbone and is larger in contour accuracy $\mathcal{F}$. Code is available at
\href{https://github.com/KotaroHorikawa/AnchorUp}{github.com/KotaroHorikawa/AnchorUp}.
\end{abstract}

%% file: sec/1_intro.tex
\section{Introduction}
\label{sec:intro}

\begin{figure}[t]
  \centering
  \includegraphics[width=\linewidth]{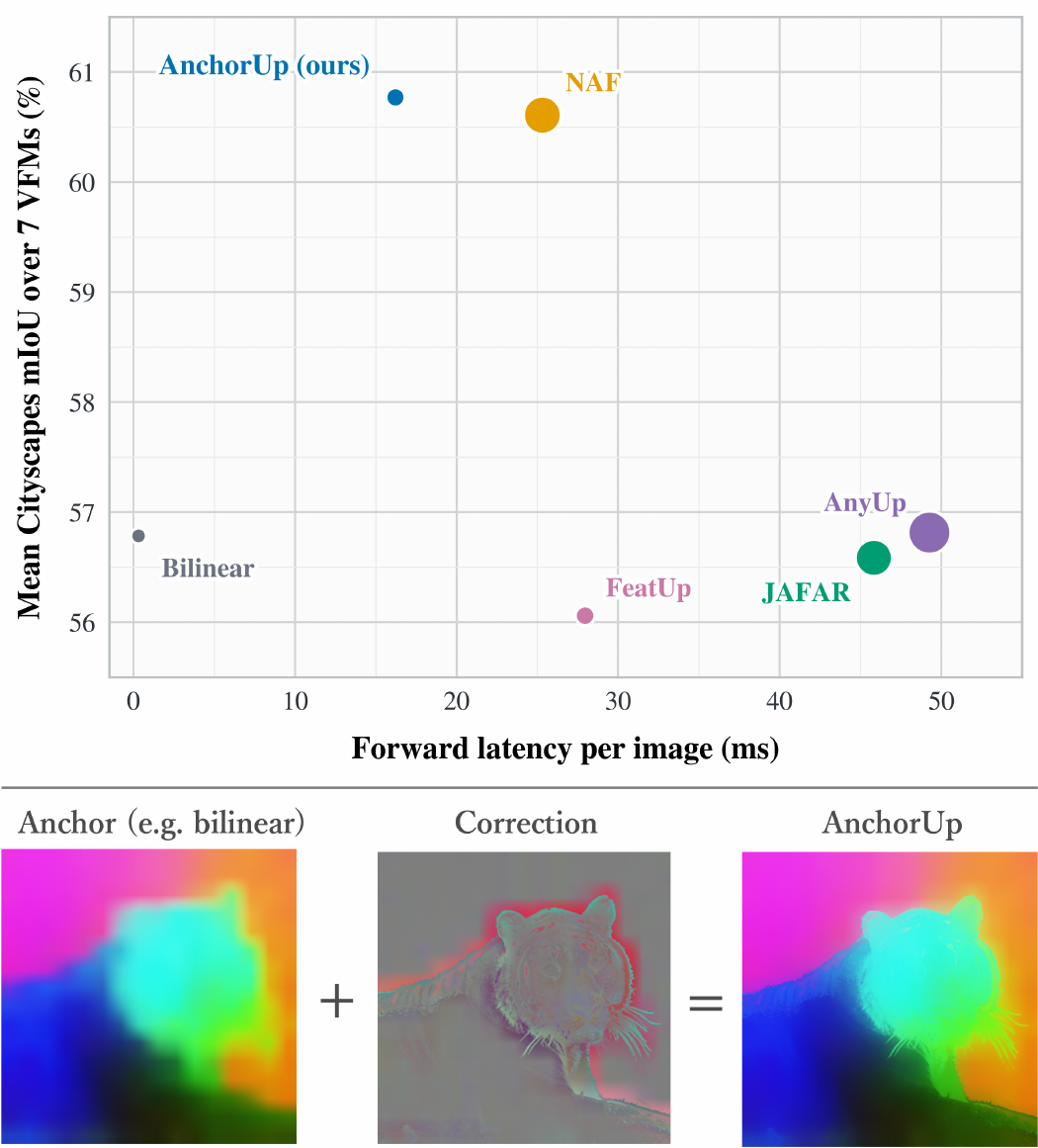}
  \caption{\textbf{Accuracy against latency} on Cityscapes averaged over seven backbones. Circle size shows the number of parameters (Tables~\ref{tab:baselines} and~\ref{tab:efficiency}). \emph{Top:} \method{} achieves a better accuracy-speed trade-off than baselines: NAF. \emph{Bottom:} Because our learned part supplies only a correction to a classical interpolation rather than the whole feature map, it focuses solely on missing details (outlines, whiskers, and fur) while remaining flat over the background. Feature maps are projected onto the first three principal components of each image; in the correction, zero appears as mid-gray.}
  
  \label{fig:teaser}
\end{figure}

Vision foundation models (VFMs) such as DINOv3~\cite{simeoni2025dinov3},
CLIP~\cite{radford2021clip}, SigLIP~2~\cite{tschannen2025siglip2} and
RADIO~\cite{heinrich2024radiov25} have become the default source of visual features,
but they emit them on a patch grid that is 14 or 16 times coarser than the image.
Segmentation, tracking and depth all need those features at every pixel, and
feature upsampling exists to close that gap.

Recent work on this problem
\cite{fu2024featup,couairon2025jafar,huang2025loftup,wimmer2025anyup,chambon2026naf}
treats it as regression: from the low-resolution features $\Flr$ and the image $I$,
the goal is to recover the high-resolution features $\Fhr$ that the VFM would have produced
at full resolution. Each model differs in how they use the image as guidance, but they
all predict the same thing: the whole high-resolution feature map, $\Fhat$.

How much of this prediction needs learning? We hypothesize that most of it does
not. Feature maps of natural images are smooth over most of their area, so an
interpolation should reproduce most of each map; what remains is mostly edges and
fine detail.
NAF's own comparison supports this: across four datasets with a fixed backbone
(Table~3 in~\cite{chambon2026naf}), JAFAR, FeatUp and AnyUp fail to outperform
bicubic on average. If the hypothesis is correct, an upsampler that outputs $\Fhat$
directly must still produce all of it, so much of its capacity goes to what classical
interpolation already provides.

We therefore change what the learned part of the model is asked for. \method{}
keeps a classical upsampler, bilinear or bicubic with no parameters, as an
\emph{anchor} $\Fanchor$ and takes a second prediction $\Flearn$ from a small NAF-style
upsampler. The NAF-style upsampler has parameters in one place only: a lightweight
encoder of the input image, which we call the \emph{guidance network}. We combine the
two as
\begin{equation}
  \Fhat = \Fanchor + \alpha\,\bigl(\Flearn - \Fanchor\bigr),
    \label{eq:anchorup}
\end{equation}
where $\alpha = \sigma(\gamma)$ and $\gamma$ is a single learnable scalar. The
NAF-style upsampler still predicts the whole high-resolution feature map, but only its
difference from
the anchor enters the output, scaled by $\alpha$; the sigmoid keeps $\alpha$ in
$(0,1)$, so the output always lies between the two estimates. 
$\alpha = 0$ means that the
output is the interpolant itself; training starts at $\alpha = 0.5$ and moves the gate
where the loss wants it. Because the anchor already gets the smooth regions right, the
training signal reaches the network almost only where the anchor is wrong: at edges
and fine detail. This is what we find: the correction $\alpha(\Flearn - \Fanchor)$
is high-pass in the spectrum and lives on object boundaries in image space as shown in \cref{fig:teaser}
(Details are in \cref{sec:analysis}). Namely, by relieving the network from reproducing low-frequency content, it can concentrate its capacity entirely on high-frequency components, such as sharp edges and fine details. This is why the guidance network can be small.

Overall, our contributions are as follows:
\begin{itemize}
\itemsep0.15em
\item We propose \method{}, a feature upsampler that
      learns only a correction to a classical interpolation.
\item \method{} is on par with NAF on Pascal VOC and Cityscapes and beats it on
      DAVIS-2017 on every backbone tested, at a quarter NAF's parameters and 1.6 times faster than NAF, with only the first of two training stages of NAF.
\item We show what the learned part does when an interpolation is given: its correction
      is high-pass in the spectrum and lives on object boundaries in image space, so
      the guidance network can be small.
\end{itemize}

%% file: sec/2_related.tex
\section{Related Work}
\label{sec:related}

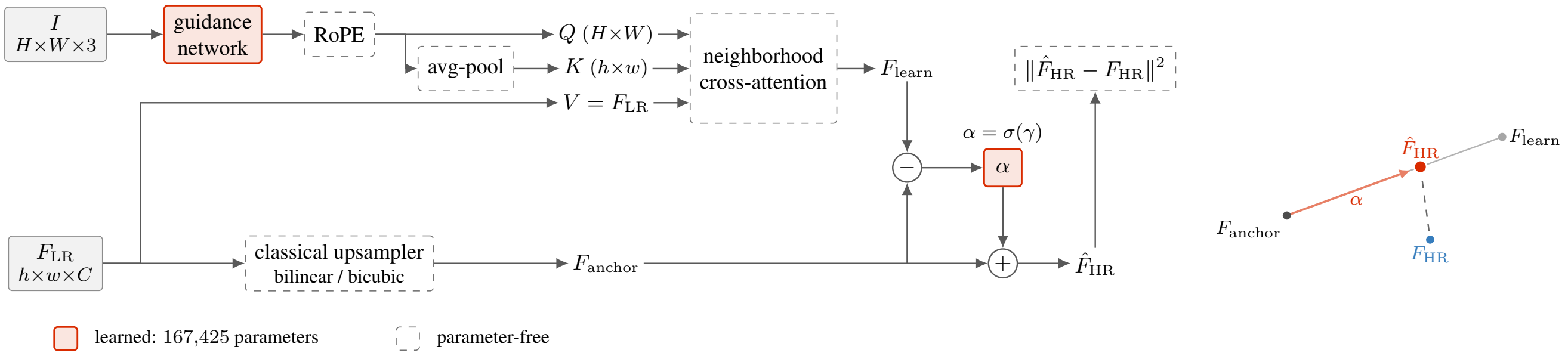
\begin{figure*}[t]
  \centering
  \resizebox{\textwidth}{!}{\input{figures/architecture}}
  \caption{\textbf{\method{}.} (a) The low-resolution feature map is upsampled in two
  ways: by a classical upsampler with no parameters, which gives the anchor, and by a
  NAF-style upsampler, a guidance network followed by neighborhood cross-attention. In
  the attention, queries come from the encoded image, keys from its average pooling, and
  values from the feature map itself. Only the guidance network and the gate
  $\alpha=\sigma(\gamma)$ have learnable parameters; the gate controls how far the output moves from
  the anchor toward the NAF-style prediction. (b) The output is a point on the segment
  between the two predictions: the network picks the direction, the gate picks the
  distance. The dashed line is the remaining error to the target.}
  \label{fig:architecture}
\end{figure*}

\paragraph{Feature upsampling.}
Feature upsamplers split into two families: training-free interpolation and
learned upsamplers. Interpolation, such as bilinear or bicubic, is simple and works
with any backbone, but it cannot recover the detail that the backbone's
downsampling removed. Learned upsamplers were introduced to recover it from
the image. Decoders for dense prediction, such as segmentation, did this
earlier, training the upsampler end-to-end inside the task
network~\cite{wang2019carafe,lu2022sapa,lu2022fade,liu2023dysample}; we focus
on the VFM setting, where the upsampler is trained apart from any downstream
task. FeatUp~\cite{fu2024featup} introduced the task for VFM features, with a
stack of joint bilateral filters~\cite{kopf2007jbu} or a per-image implicit
field. A convolutional upsampler is restricted to fixed scaling factors, so
recent methods use cross-attention
instead~\cite{couairon2025jafar,huang2025loftup}: at each output pixel, a
query built from the image is matched against keys, and the match decides how
the low-resolution features are mixed. Queries can be built for any output
grid, so one trained model upsamples to any target resolution.

These cross-attention methods differ in what builds the keys and in whether
one trained model transfers across backbones. JAFAR~\cite{couairon2025jafar} fuses image and VFM features into its keys;
LoftUp~\cite{huang2025loftup} uses the VFM features directly and trains against
pseudo-ground-truth built from SAM masks and self-distillation; both must be
retrained for every backbone. To remove this retraining,
AnyUp~\cite{wimmer2025anyup} builds its keys like JAFAR from the image and
the VFM features, but first maps the VFM features to a fixed dimension: the
keys no longer depend on the backbone's feature dimension, so one trained
model serves backbones it never saw.
NAF~\cite{chambon2026naf} goes further in the same direction: to depend on
the backbone as little as possible, it builds queries and keys from a single
image encoder without the VFM features, and keeps attention local for speed.
It is the fastest of these models, and zero-shot it is the most accurate.
UPLiFT~\cite{walmer2026uplift} uses iterative local attention for efficiency, but
again needs one model per backbone. These methods suggest that transfer
improves as the model depends less on the backbone's features.

\method{} combines a training-free interpolation with a learned upsampler. The training-free
interpolant stays in the output as the anchor, and the learned part, NAF's
upsampler unchanged (Section~\ref{sec:method-learned}), supplies only the
correction: it no longer has to reproduce what interpolation already provides.

\paragraph{Residual prediction.}

When a simple estimate of the output exists, predicting the difference from it is
a natural idea, %
explored across diverse domains.
Residual networks~\cite{he2016resnet}
built an architecture on it. Super-resolution used it on the output itself:
VDSR~\cite{kim2016vdsr} regressed the residual over a bicubic upsample rather
than the image~\cite{dong2014srcnn}. 
Classical interpolation appears in the feature upsampling literature only as a baseline to be
replaced, blurry and blind to content~\cite{couairon2025jafar}, and so every
upsampler predicts the whole high-resolution feature map. Yet bilinear and
bicubic are often not beaten (Table~3 of~\cite{chambon2026naf}), suggesting that most of the map does not need learning. Our \method{} keeps the interpolant in
the output and learns only the difference from it.

\paragraph{Zero-shot transfer.}
A separate upsampler per backbone is impractical, so recent work evaluates
zero-shot transfer: train once, apply to backbones never seen in
training~\cite{wimmer2025anyup,chambon2026naf}. We adopt this protocol and
expect the anchor to help transfer in particular: it is the same
parameter-free operation for every backbone.

%% file: figures/architecture.tex
\colorlet{learnred}{red!60!orange!85!black}
\begin{tikzpicture}[
  font=\small,
  blob/.style   = {draw=black!45, fill=black!5, rounded corners=1.5pt,
                   minimum height=6.5mm, inner xsep=4pt, align=center},
  learned/.style= {draw=learnred, fill=learnred!12, line width=0.7pt,
                   rounded corners=1.5pt, minimum height=6.5mm, inner xsep=4pt, align=center},
  frozen/.style = {draw=black!55, fill=white, dashed, rounded corners=1.5pt,
                   minimum height=6.5mm, inner xsep=4pt, align=center},
  tensor/.style = {align=center, inner sep=1.5pt, fill=white},
  op/.style     = {circle, draw=black!65, inner sep=0.9pt, line width=0.6pt, fill=white},
  arr/.style    = {-{Latex[length=1.8mm,width=1.6mm]}, draw=black!65, line width=0.6pt},
  plain/.style  = {draw=black!65, line width=0.6pt},
]

\begin{scope}
  \node[blob] (img) at (0,  1.60) {$I$\\[-2pt]\footnotesize $H{\times}W{\times}3$};
  \node[blob] (flr) at (0, -1.75) {$F_{\mathrm{LR}}$\\[-2pt]\footnotesize $h{\times}w{\times}C$};

  \node[learned] (enc)  at (2.30, 1.60) {guidance\\network};
  \node[frozen]  (rope) at (4.15, 1.60) {RoPE};
  \node[frozen]  (pool) at (6.00, 1.10) {avg-pool};

  \node[tensor]  (q) at (8.05, 1.60) {$Q$ \footnotesize$(H{\times}W)$};
  \node[tensor]  (k) at (8.05, 1.10) {$K$ \footnotesize$(h{\times}w)$};
  \node[tensor]  (v) at (8.05, 0.60) {$V = F_{\mathrm{LR}}$};

  \node[frozen, minimum height=16mm] (attn) at (10.35, 1.10) {neighborhood\\cross-attention};
  \node[tensor] (lout) at (12.45, 1.10) {$F_{\mathrm{learn}}$};

  \node[frozen] (anc)  at (4.15, -1.75) {classical upsampler\\[-1pt]\footnotesize bilinear / bicubic};
  \node[tensor] (aout) at (8.05, -1.75) {$F_{\mathrm{anchor}}$};

  \node[op]     (minus) at (12.45, -0.35) {$-$};
  \node[learned, minimum height=5.5mm, minimum width=5.5mm] (gate) at (13.85, -0.35) {$\alpha$};
  \node[op]     (plus)  at (13.85, -1.75) {$+$};
  \node[tensor] (out)   at (15.20, -1.75) {$\hat F_{\mathrm{HR}}$};
  \node[frozen] (loss)  at (15.20, 1.10) {$\lVert \hat F_{\mathrm{HR}} - F_{\mathrm{HR}}\rVert^2$};

  \draw[arr] (img)  -- (enc);
  \draw[arr] (enc)  -- (rope);
  \draw[arr] (rope) -- (q);
  \draw[arr] (rope.east) -- ++(0.45,0) |- (pool.west);
  \draw[arr] (pool) -- (k);

  \draw[arr] (flr.east) -- ++(0.55,0) |- (v.west);

  \draw[arr] (q.east) -- (attn.west |- q);
  \draw[arr] (k.east) -- (attn.west |- k);
  \draw[arr] (v.east) -- (attn.west |- v);
  \draw[arr] (attn) -- (lout);

  \draw[arr] (flr) -- (anc);
  \draw[arr] (anc) -- (aout);
  \draw[arr] (aout.east) -- (plus.west);
  \draw[arr] (minus |- aout) -- (minus.south);

  \draw[arr] (lout) -- (minus);
  \draw[arr] (minus) -- (gate);
  \draw[arr] (gate) -- (plus);
  \draw[arr] (plus) -- (out);
  \draw[arr] (out.north) -- (loss.south);

  \node[anchor=south, font=\footnotesize, inner sep=2pt] at (gate.north) {$\alpha=\sigma(\gamma)$};

  \node[learned, minimum height=3.4mm, minimum width=3.4mm, inner sep=0pt] at (0.15, -2.85) {};
  \node[anchor=west, font=\footnotesize] at (0.45, -2.85) {learned: $167{,}425$ parameters};
  \node[frozen, minimum height=3.4mm, minimum width=3.4mm, inner sep=0pt] at (5.15, -2.85) {};
  \node[anchor=west, font=\footnotesize] at (5.45, -2.85) {parameter-free};
  \node[anchor=base, font=\footnotesize\bfseries] at (7.60, -3.55)
       {(a) AnchorUp architecture.};
\end{scope}

\begin{scope}[xshift=18.0cm, yshift=-1.05cm]
  \coordinate (A) at (0, 0);
  \coordinate (L) at (3.15, 1.15);
  \coordinate (P) at ($(A)!0.620!(L)$);   %
  \coordinate (T) at ($($(A)!0.551!(L)$)!1.05cm!-90:(L)$); %

  \draw[draw=black!30, line width=0.6pt] (A) -- (L);
  \draw[-{Latex[length=1.7mm,width=1.5mm]}, draw=learnred!60, line width=0.9pt] (A) -- ($(A)!0.93!(P)$);
  \draw[dashed, draw=black!60, line width=0.6pt] (P) -- (T);

  \fill[black!70] (A) circle (1.7pt);
  \fill[black!35] (L) circle (1.7pt);
  \filldraw[fill=learnred, draw=white, line width=0.7pt] (P) circle (2.6pt);
  \fill[wacvblue] (T) circle (1.7pt);

  \node[anchor=north east, font=\footnotesize, inner sep=2pt] at (A) {$F_{\mathrm{anchor}}$};
  \node[anchor=west,  font=\footnotesize, inner sep=2.5pt] at (L) {$F_{\mathrm{learn}}$};
  \node[anchor=south, font=\footnotesize, inner sep=3.5pt, text=learnred] at (P) {$\hat F_{\mathrm{HR}}$};
  \node[anchor=north, font=\footnotesize, inner sep=2.5pt, text=wacvblue] at (T) {$F_{\mathrm{HR}}$};
  \node[anchor=north west, font=\footnotesize, inner sep=1pt, text=learnred] at ($(A)!0.28!(L)$) {$\alpha$};

  \node[anchor=base, font=\footnotesize\bfseries] at (1.60, -2.50)
       {(b) The output as a point on a segment.};
\end{scope}

\end{tikzpicture}

%% file: sec/3_method.tex
\section{Method}
\label{sec:method}

\subsection{Problem setting}

A frozen backbone turns an image $I \in \mathbb{R}^{H \times W \times 3}$ into a
feature map $\Flr \in \mathbb{R}^{h \times w \times C}$ on a patch grid, with
$h = H/p$ for patch size $p$. The target $\Fhr$ is the high-resolution feature map that
the same backbone produces at full resolution. An upsampler takes $(I, \Flr)$ and
predicts $\Fhat$ in \cref{eq:anchorup}, and is trained to minimize $\lVert \Fhat - \Fhr \rVert^2$. At test time, it is applied to backbones that was not trained on in order to show the generalization ability. The target $\Fhr$ has more
detail than $\Flr$, and that detail has to come from the image.

\subsection{\method{}}
\label{sec:method-anchorup}

Let $u$ be a classical upsampler, bilinear or bicubic, applied to each channel, and
let $\Fanchor = u(\Flr)$. Let $\Flearn = u_\theta(I, \Flr)$ be the prediction of a
learned upsampler. In this paper, $u_\theta$ is NAF's upsampler with a smaller guidance dimension
(Section~\ref{sec:method-learned}). \method{} predicts
\begin{equation*}
  \Fhat \;=\; \Fanchor + \alpha \bigl( \Flearn - \Fanchor \bigr),
\end{equation*}
where $\alpha = \sigma(\gamma)$, $\gamma \in \mathbb{R}$ is a single learnable
scalar, and $\sigma$ is the logistic function.

\cref{eq:anchorup} changes what $u_\theta$ has to learn. Existing learned
upsamplers output $\Fhat$ directly, so they have to produce every feature, including
the regions where classical interpolation is already enough. 
In contrast, \method{} lets the anchor handle regions where classical
interpolation is already sufficient, allowing $u_\theta$ to focus primarily on
correcting the residual errors.
We hypothesize that this is easier to learn than the high-resolution feature map directly, and Section~\ref{sec:analysis} tests the hypothesis.

The gradient that reaches $\theta$ is
\begin{equation}
  \frac{\partial}{\partial \theta} \lVert \Fhat - \Fhr \rVert^2
  = 2\alpha \left( \Fhat - \Fhr \right)^{\!\top} \frac{\partial \Flearn}{\partial \theta},
  \label{eq:gradient}
\end{equation}
Thus, from the first step of training, the gradient received by $u_\theta$ is
scaled by both $\alpha$ and the current output error. As regions already well
explained by the anchor become accurate, their gradients shrink, concentrating
learning on interpolation failures, which often occur at boundaries and fine
structures.
This is the mechanism the paper rests on.

Because $\alpha \in (0,1)$, the output lies on the segment between the two
predictions (Figure~\ref{fig:architecture}b). At $\alpha = 0$, it is the interpolation, and it is the NAF-style upsampler alone at $\alpha = 1$. We initialize $\gamma = 0$, so
training starts at the midpoint $\alpha = 0.5$ and is free to move either way.
Our method adds only one parameter, $\gamma$, to $u_\theta$. The forward pass is
given as Algorithm~1 in the supplementary material.

\subsection{NAF-style upsampler}
\label{sec:method-learned}

We take $u_\theta$ from NAF~\cite{chambon2026naf} without modification: it is NAF's
upsampler itself, made of a guidance network 
that encodes the image, and neighborhood
cross-attention that uses this encoding to weight the nearby low-resolution features
at each output pixel. The guidance network, NAF's guidance encoder, turns $I$ into a
$d$-channel encoding with two parallel convolutional stacks, one with $1{\times}1$
kernels and one with $3{\times}3$, each with $d/2$ channels, followed by rotary position
embedding (RoPE)~\cite{su2021roformer,heo2024rope}. The queries are this encoding at
the output resolution $H \times W$; the keys are the same encoding average-pooled onto
the $h \times w$ grid of $\Flr$; the values are $\Flr$ itself. Attention is restricted
to a $k \times k$ neighborhood~\cite{ramachandran2019sasa}, so the cost is
$O(HW k^2)$, linear in the number of output pixels, instead of $O(HW\,hw)$ for
attention over the whole grid.

Within $u_\theta$, only the guidance network has parameters. The position embedding
and the attention have none, and the values are not projected. So nothing in
$u_\theta$ depends on the channel count $C$, and the same weights work on any
backbone. The parameter count is close to quadratic in $d$: 662{,}528 at $d = 256$
and 167{,}424 at $d = 128$. Original NAF uses $d = 256$. Our hypothesis says the guidance
network can be smaller
because it can concentrate its capacity entirely on only high-frequency components, so we set $d = 128$ and use it throughout; with $\gamma$,
\method{} has 167{,}425 parameters, a quarter of NAF's.

\subsection{Scalar gate}
\label{sec:method-gate}

The gate $\alpha$ controls how far the output moves from the anchor toward
$u_\theta$'s prediction. We keep $\alpha$ a single number, shared by every pixel and
channel. \method{} then has the parameters of $u_\theta$ plus one, so its comparison
with $u_\theta$ trained on its own (Tables~\ref{tab:main} and~\ref{tab:davis}) is
fair: any gain comes from the anchor, not from extra capacity.

At inference, the model is frozen, so $\Fanchor$ and $\Flearn$ are fixed, and the loss
$\lVert \Fhat - \Fhr \rVert^2$ is then a quadratic in $\alpha$ alone. Its minimizer is
\begin{equation}
  \alpha^\star
  = \frac{\bigl( \Fhr - \Fanchor \bigr)^{\!\top} \bigl( \Flearn - \Fanchor \bigr)}
         {\bigl\lVert \Flearn - \Fanchor \bigr\rVert^2}.
  \label{eq:alpha-star}
\end{equation}
The optimum $\alpha^\star$ is the gate the loss would pick if it could see the target.
It requires $\Fhr$ and thus cannot be used at test time. During actual
inference, we instead use $\alpha = \sigma(\gamma)$, which is learned during
training and kept fixed: $0.620$ with the bilinear anchor and $0.771$ with the
bicubic anchor. Each anchor settles at its own mixture, so learning $\gamma$
--- a single parameter --- removes a choice that would otherwise require
tuning for each anchor.
However, on images set aside from
training, we can run the backbone at full resolution, build $\Fhr$, and solve
Eq.~\ref{eq:alpha-star} directly, with no label needed. In this case, $\alpha^\star$
measures how close the anchor already is to the target
(supplementary material).

\subsection{Training}
\label{sec:method-training}

We follow the first stage of NAF's recipe, using the configuration from its released code~\cite{chambon2026naf}. Training runs on
ImageNet~\cite{deng2009imagenet} with features from
DINOv3-B/16~\cite{simeoni2025dinov3}: targets are features of $512{\times}512$ images,
inputs are features of the same images downsampled by 2, batch size 4,
AdamW~\cite{loshchilov2019adamw} at $2\times10^{-4}$ with weight decay $10^{-5}$. The
neighborhood kernel size is $k = 9$, with four attention heads; the remaining
settings also follow NAF. The gate $\gamma$ is trained jointly with $\theta$; no
schedule, loss term or hyperparameter is added for it. We train for 25{,}000 steps,
the length of NAF's first stage, and report that checkpoint.

NAF uses $d = 256$ (Section~\ref{sec:method-learned}) and is trained in two stages;
we train \method{} using only the first stage. The second stage adapts the model to resolution
changes, with 2{,}500 more steps, $1024{\times}1024$ targets and inputs from
resolutions between 256 and 896, and is worth $+0.4$ mIoU on VOC linear probing
for NAF at $d = 256$ (Appendix~B.1 of~\cite{chambon2026naf}). Our evaluations upsample features to the full resolution of $448 \times 448$ images for linear probing and by a factor of two on DAVIS-2017 frames. In these evaluations, \method{} trained without the second stage is on par with NAF trained through both stages on VOC and Cityscapes, and beats it on DAVIS-2017 (Section~\ref{sec:experiments}). By Eq.~\ref{eq:anchorup}, an error in $u_\theta$ at a resolution unseen during training enters the output scaled by $\alpha$, while the anchor itself is computed by the same rule at every resolution (supplementary material).

To measure what the anchor itself contributes, we also train NAF at $d = 128$
(Section~\ref{sec:method-learned}) with exactly this recipe: the same model as
\method{}, with no anchor and no gate. Tables~\ref{tab:main} and~\ref{tab:davis}
report this control as the third NAF row.

%% file: sec/4_experiments.tex
\section{Experiments}
\label{sec:experiments}

\input{tables/tab_main}
\input{tables/tab_baselines}
\input{tables/tab_davis}
\input{tables/tab_efficiency}
\input{tables/tab_seeds}

NAF and \method{} are trained on DINOv3-B/16~\cite{simeoni2025dinov3} alone and
applied unchanged to six other backbones, all at base size. The tables group the
seven by pretraining objective: image-only (DINOv3,
Franca~\cite{venkataramanan2025franca}), image-text (CLIP~\cite{radford2021clip},
SigLIP~2~\cite{tschannen2025siglip2}, and PEcore, the core model of Perception
Encoder~\cite{bolya2025pe}), and distilled from teacher models (PEspatial, its
spatial variant, and RADIO~\cite{ranzinger2024amradio,heinrich2024radiov25}).
We retrain every learned upsampler of Table~\ref{tab:baselines} ourselves under its
published configuration; FeatUp and JAFAR require one model per backbone. The
supplementary material provides the full details of the backbones, protocols, and training configuration.

\subsection{Semantic segmentation}
\label{sec:crossbackbone}

\paragraph{Setup.}
We use NAF's evaluation pipeline without modification~\cite{chambon2026naf}: with the backbone and upsampler frozen, we train a linear layer on the upsampled features and report the mIoU at epoch 20 on Pascal VOC~\cite{everingham2010voc} and Cityscapes~\cite{cordts2016cityscapes}.

\paragraph{Across backbones.}
Two comparisons matter in Table~\ref{tab:main}. The first is against the control:
NAF at $d = 128$ trained with our recipe (Section~\ref{sec:method-training}), so
that the two models differ only by Eq.~\ref{eq:anchorup}. \method{} (bilinear)
gains $0.56$ mIoU on the Cityscapes mean and leads on five of the seven backbones.
On VOC, \method{} (bilinear) ends $0.18$ below the third NAF row on the mean, 
and the
deficit has a consistent pattern: it wins on the four image-only and distilled
backbones and loses on the three image--text ones, by up to $1.05$ mIoU.
\method{} (bicubic) turns those losses into gains and ends $0.36$ above the third NAF row.
The anchor's kernel decides which backbones gain; we return to it below. Across three training seeds, \method{} (bilinear) stays below the control on the VOC mean and above it on the Cityscapes mean (Table~\ref{tab:seeds}). The second comparison is against NAF as published, at $d = 256$ with both
training stages: \method{} (bilinear) ties it on VOC and is $0.16$ ahead of it on
Cityscapes, with a quarter of the parameters and one training stage.

\paragraph{Anchor kernel.}
The choice of the anchor matters: \method{} (bicubic) is $0.54$ mIoU better than
\method{} (bilinear) on the VOC mean, and $0.42$ worse on Cityscapes
(Table~\ref{tab:main}). The two kernels differ in a classical way: bicubic has
negative weights at the edges of its window (side lobes), so it renders edges
sharper but overshoots around them; bilinear has no negative weights and only
blurs. We interpret the split as the sharpening paying off on VOC's large objects and the
overshoot hurting on Cityscapes' thin structures.

\begin{figure*}[t]
  \centering
  \begin{minipage}{\textwidth}
    \centering
    \newcommand{\pcahead}[1]{\makebox[\dimexpr(\linewidth-2pt)/6\relax][c]{\footnotesize #1}}%
    \pcahead{Image}\pcahead{LR features}\pcahead{Anchor}\pcahead{Learned}%
    \pcahead{Correction}\pcahead{\method{}}\\[1.5pt]
    \includegraphics[width=\linewidth]{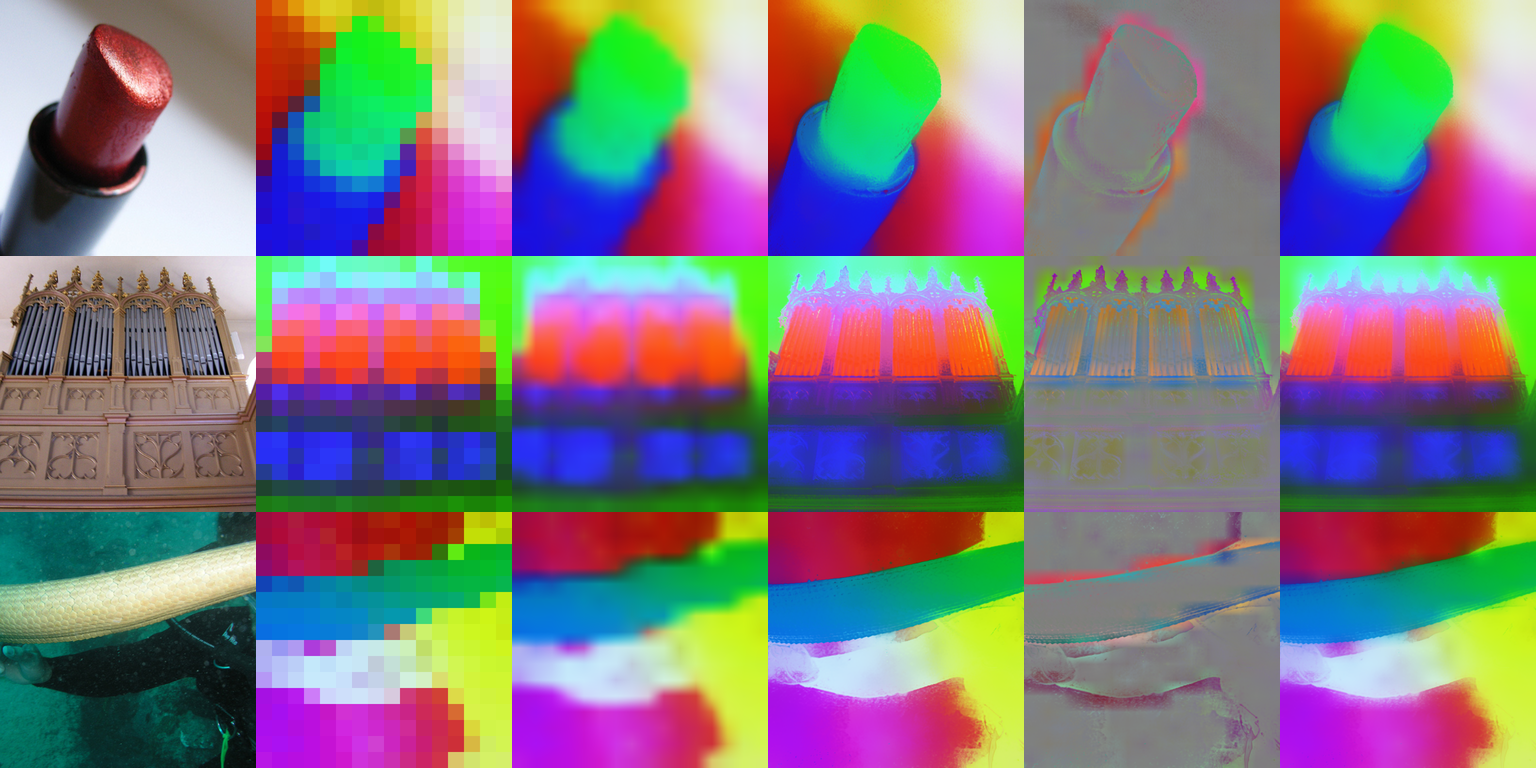}
  \end{minipage}
  \caption{\textbf{The correction is a boundary signal.} Columns: input,
  low-resolution features, anchor $\Fanchor$, learned estimate $\Flearn$,
  correction $C_\theta = \alpha(\Flearn - \Fanchor)$, and \method{} output
  $\Fhat$. Feature maps are projected onto the first three principal components of
  each image and shown as RGB; in the correction column, zero appears as
  mid-gray. The
  correction is flat inside regions and concentrates on contours and on the fine
  repeated structure the anchor cannot resolve (rows 2 and 3). Added to the
  anchor (third column), this is the detail the output (last column) recovers.}
  \label{fig:pca}
\end{figure*}

\begin{figure*}[t]
  \centering
  \begin{minipage}[c]{0.5\textwidth}
    \centering
    \includegraphics[width=\linewidth]{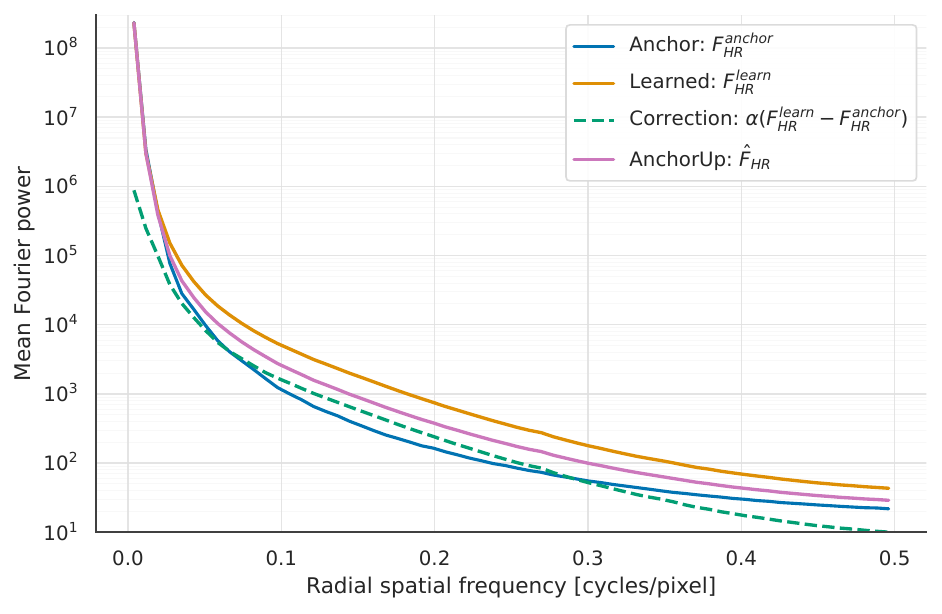}
  \end{minipage}\hfill
  \begin{minipage}[c]{\columnwidth}
    \caption{\textbf{The correction is high-pass.} Radially averaged Fourier power
  of each term of Eq.~\ref{eq:anchorup}: the anchor $\Fanchor$, the learned
  estimate $\Flearn$, the correction $C_\theta = \alpha(\Flearn - \Fanchor)$,
  and the output $\Fhat$. Two readings matter. At the lowest frequencies the
  correction carries about two orders of magnitude less power than the anchor
  ($10^6$ against $2 \times 10^8$): the smooth content of the output comes from
  the anchor, not from the network. Between roughly $0.06$ and $0.3$ cycles per
  pixel the correction carries more power than the anchor: the learned part
  spends its output where the interpolation cannot reach. The measurement
  protocol is in the supplementary material.}
    \label{fig:spectrum}
  \end{minipage}
\end{figure*}

The gains from the sharper kernel track pretraining rather than accuracy: on both
benchmarks, %
the three main differences between \method{} (bicubic) and \method{} (bilinear) are on the three image--text backbones. We explain this through what the
pretraining asks of a patch. CLIP and PEcore are supervised through a single
image-level embedding, and SigLIP~2 adds a patch-level loss only late in
pretraining~\cite{tschannen2025siglip2}; features trained this way place
boundaries less precisely, so the sharpening adds contrast they lack, and the
overshoot finds little precise structure to corrupt. The Perception Encoder pair
tests this interpretation directly: PEspatial distills spatial correspondence into
PEcore~\cite{bolya2025pe}, and the gain PEcore shows is gone on PEspatial. 
We keep bilinear as the default: it is on par with or above the published NAF
on both benchmarks.

\paragraph{Against other upsamplers.}
On Cityscapes, FeatUp, JAFAR, and AnyUp all beat bilinear on the three image--text backbones, and all lose to it on DINOv3, PEspatial, and RADIO (Table~\ref{tab:baselines}). These three are the backbones whose pretraining supervises the patch grid itself: Gram anchoring in DINOv3~\cite{simeoni2025dinov3}, distilled spatial correspondence in
PEspatial~\cite{bolya2025pe}, and distillation toward the teachers' spatial fidelity
in RADIO~\cite{ranzinger2024amradio}. A grid supervised this way varies
coherently from patch to patch, so interpolating between patches already comes
close to the high-resolution features: bilinear thus becomes a strong baseline. In
\method{}, bilinear is one term of Eq.~\ref{eq:anchorup} and the learned part only
adds a correction; \method{} is above bilinear on all seven backbones.

\subsection{Video object segmentation}
\label{sec:davis}

DAVIS-2017~\cite{pont2017davis} provides only the first-frame mask, and we propagate labels
to later frames based on the similarity of the upsampled features. Nothing is trained on
video, so the score measures the spatial correspondence carried by the features. We report
region similarity $\mathcal{J}$, contour accuracy $\mathcal{F}$, and their mean over
six backbones, excluding PEcore.

\method{} (bilinear) outperforms the two-stage NAF model across all six backbones in
Table~\ref{tab:davis}, five of which are unseen during training. Compared to NAF at $d = 128$
trained with our recipe, the margin is $2.49$ on the mean
$\mathcal{J}\&\mathcal{F}$, again across every backbone. Across three training seeds, \method{} consistently outperforms the control (Table~\ref{tab:seeds}).
The margins are wider than under linear probing because DAVIS uses the features
directly: no head is trained that could absorb feature defects, and each pixel
takes the labels of its nearest neighbors in feature space within a local window
over preceding frames. For an interior pixel, nearly all neighbors carry the
same label, so even a blurred feature votes correctly. At a boundary, a blurred
feature votes for the wrong side, and the wrong label is propagated to subsequent frames.
DAVIS therefore amplifies boundary-level feature quality (NAF's second stage is worth $1.66$ here,
versus $0.27$ on Cityscapes), and boundaries are where the correction lives
(Section~\ref{sec:analysis}): the gain is larger on contour accuracy $\mathcal{F}$
than on region similarity $\mathcal{J}$, including in the controlled comparison.

\subsection{Cost}
\label{sec:cost}

We time the upsampler alone, for one pass from $28 \times 28$ features to $448 \times 448$ output on a single RTX PRO 6000 Blackwell GPU. \method{} is $1.6\times$ faster than NAF, with a quarter of its parameters, and holds the smallest peak memory of the learned upsamplers in Table~\ref{tab:efficiency}. Switching the anchor from bilinear to bicubic adds $0.5$\,ms, the same gap that separates bilinear from bicubic when run on their own (Table~\ref{tab:efficiency}); the anchor costs nothing beyond the interpolation itself. The supplementary material gives the measurement setup, the anchor cost, and the guidance-dimension sweep that fixes $d = 128$.

%% file: tables/tab_main.tex
\begin{table*}[t]
  \centering
  \small
  \setlength{\tabcolsep}{4pt}
  \begin{tabular}{l l c r c *{7}{c} c c}
    \toprule
    & & & & & \multicolumn{2}{c}{Images only} & \multicolumn{3}{c}{Images $+$ text}
      & \multicolumn{2}{c}{Distilled} & & \\
    \cmidrule(lr){6-7}\cmidrule(lr){8-10}\cmidrule(lr){11-12}
    Method & Anchor & $d$ & \#Params & Stage 2
      & DINOv3 & Franca & CLIP & SigLIP2 & PEcore & PEspatial & RADIO & Mean & $\Delta$ \\
    \midrule
    \multicolumn{14}{l}{\textit{(a) Pascal VOC}} \\
    NAF~\cite{chambon2026naf} & -- & 256 & 662{,}528 & \checkmark & \underline{87.81} & \textbf{84.30} & 80.25 & 77.83 & 72.61 & 78.04 & \underline{86.61} & 81.06 & -- \\
    NAF & -- & 128 & 167{,}424 & \checkmark & \textbf{87.83} & 84.21 & 79.98 & 77.40 & 72.16 & \underline{78.16} & 86.51 & 80.89 & \dneg{0.17} \\
    NAF & -- & 128 & 167{,}424 & & 87.34 & 84.24 & \underline{80.53} & \underline{78.85} & \underline{73.62} & 77.60 & 86.55 & \underline{81.24} & \dpos{0.18} \\
    \textbf{AnchorUp} & bilinear & 128 & 167{,}425 & & 87.71 & \underline{84.27} & 80.23 & 77.80 & 72.63 & \textbf{78.18} & \underline{86.61} & 81.06 & 0.00 \\
    \textbf{AnchorUp} & bicubic & 128 & 167{,}425 & & 87.30 & 84.20 & \textbf{80.78} & \textbf{79.53} & \textbf{74.61} & 78.15 & \textbf{86.64} & \textbf{81.60} & \dpos{0.54} \\
    \midrule
    \multicolumn{14}{l}{\textit{(b) Cityscapes}} \\
    NAF~\cite{chambon2026naf} & -- & 256 & 662{,}528 & \checkmark & 66.29 & \underline{63.80} & \underline{57.04} & 58.14 & 55.10 & \underline{61.35} & \underline{62.55} & \underline{60.61} & -- \\
    NAF & -- & 128 & 167{,}424 & \checkmark & \underline{66.39} & 63.71 & 56.85 & 57.77 & 54.80 & 61.31 & 62.53 & 60.48 & \dneg{0.13} \\
    NAF & -- & 128 & 167{,}424 & & 65.37 & 63.22 & 56.50 & \underline{58.28} & \underline{55.41} & 60.72 & 61.98 & 60.21 & \dneg{0.40} \\
    \textbf{AnchorUp} & bilinear & 128 & 167{,}425 & & \textbf{66.45} & \textbf{63.99} & \textbf{57.07} & \underline{58.28} & 55.25 & \textbf{61.61} & \textbf{62.73} & \textbf{60.77} & \dpos{0.16} \\
    \textbf{AnchorUp} & bicubic & 128 & 167{,}425 & & 65.53 & 63.48 & 56.78 & \textbf{58.58} & \textbf{55.67} & 60.82 & 61.61 & 60.35 & \dneg{0.26} \\
    \bottomrule
  \end{tabular}
  \caption{\textbf{Cross-backbone linear probing.}  All upsamplers are trained
  once on DINOv3-B/16 and applied unchanged to the other six backbones.
  \emph{Anchor} %
  is the interpolation used in Eq.~\ref{eq:anchorup} (``--'' for NAF,
  which has none), $d$ is the guidance dimension
  (Section~\ref{sec:method-learned}),
  \emph{\#Params} the number of learnable parameters,
  and a check under \emph{Stage 2} marks
  training through the second stage of NAF's recipe
  (Section~\ref{sec:method-training}). Rows are ordered so that each differs from
  the row above in a single factor; the third NAF row therefore differs from
  \method{} only by Eq.~\ref{eq:anchorup}. The best result per column is in bold, the second best is underlined; $\Delta$ is the difference in mean from the two-stage NAF row.}
  \label{tab:main}
\end{table*}

%% file: tables/tab_baselines.tex
\begin{table*}[t]
  \centering
  \small
  \setlength{\tabcolsep}{4pt}
  \begin{tabular}{l *{7}{c} c c}
    \toprule
    & \multicolumn{2}{c}{Images only} & \multicolumn{3}{c}{Images $+$ text}
      & \multicolumn{2}{c}{Distilled} & & \\
    \cmidrule(lr){2-3}\cmidrule(lr){4-6}\cmidrule(lr){7-8}
    Method & DINOv3 & Franca & CLIP & SigLIP2 & PEcore & PEspatial & RADIO & Mean & $\Delta$ \\
    \midrule
    \multicolumn{10}{l}{\textit{(a) Pascal VOC}} \\
    Bilinear & 86.97 & 82.12 & 76.28 & 71.94 & 66.66 & 76.62 & 84.47 & 77.87 & -- \\
    FeatUp~\cite{fu2024featup} & 86.60 & 81.68 & 77.29 & 72.17 & 66.28 & 74.28 & 82.95 & 77.32 & \dneg{0.54} \\
    JAFAR~\cite{couairon2025jafar} & 87.08 & 83.65 & 79.80 & 77.51 & \underline{74.23} & 74.97 & 85.97 & 80.46 & \dpos{2.59} \\
    AnyUp~\cite{wimmer2025anyup} & 86.56 & 83.26 & 79.76 & \underline{78.56} & 74.09 & 76.77 & 85.63 & 80.66 & \dpos{2.80} \\
    NAF~\cite{chambon2026naf} & \textbf{87.81} & \textbf{84.30} & \underline{80.25} & 77.83 & 72.61 & 78.04 & \underline{86.61} & \underline{81.06} & \dpos{3.20} \\
    \textbf{AnchorUp} (bilinear) & \underline{87.71} & \underline{84.27} & 80.23 & 77.80 & 72.63 & \textbf{78.18} & \underline{86.61} & \underline{81.06} & \dpos{3.20} \\
    \textbf{AnchorUp} (bicubic) & 87.30 & 84.20 & \textbf{80.78} & \textbf{79.53} & \textbf{74.61} & \underline{78.15} & \textbf{86.64} & \textbf{81.60} & \dpos{3.74} \\
    \midrule
    \multicolumn{10}{l}{\textit{(b) Cityscapes}} \\
    Bilinear & 64.06 & 60.30 & 52.65 & 52.98 & 49.57 & 58.38 & 59.57 & 56.79 & -- \\
    FeatUp~\cite{fu2024featup} & 63.53 & 59.41 & 52.90 & 53.30 & 49.69 & 56.67 & 56.93 & 56.06 & \dneg{0.73} \\
    JAFAR~\cite{couairon2025jafar} & 63.25 & 62.38 & 53.49 & 53.41 & 50.53 & 53.75 & 59.30 & 56.59 & \dneg{0.20} \\
    AnyUp~\cite{wimmer2025anyup} & 61.52 & 59.15 & 53.59 & 54.49 & 53.08 & 57.78 & 58.10 & 56.82 & \dpos{0.03} \\
    NAF~\cite{chambon2026naf} & \underline{66.29} & \underline{63.80} & \underline{57.04} & 58.14 & 55.10 & \underline{61.35} & \underline{62.55} & \underline{60.61} & \dpos{3.82} \\
    \textbf{AnchorUp} (bilinear) & \textbf{66.45} & \textbf{63.99} & \textbf{57.07} & \underline{58.28} & \underline{55.25} & \textbf{61.61} & \textbf{62.73} & \textbf{60.77} & \dpos{3.98} \\
    \textbf{AnchorUp} (bicubic) & 65.53 & 63.48 & 56.78 & \textbf{58.58} & \textbf{55.67} & 60.82 & 61.61 & 60.35 & \dpos{3.57} \\
    \bottomrule
  \end{tabular}
  \caption{\textbf{Comparison with other feature upsamplers.}
  Every method is
  trained here under its own published configuration. NAF and AnchorUp are trained
  once on DINOv3-B/16 and applied unchanged; FeatUp and JAFAR require one model
  per backbone (Section~\ref{sec:experiments}). The NAF and AnchorUp rows are the
  models of Table~\ref{tab:main}. Best per column in bold, second best underlined;
  $\Delta$ is the difference in mean from bilinear.
  }
  \label{tab:baselines}
\end{table*}

%% file: tables/tab_davis.tex
\begin{table*}[t]
  \centering
  \small
  \setlength{\tabcolsep}{5pt}
  \begin{tabular}{l l c c *{6}{c} c c}
    \toprule
    & & & & \multicolumn{2}{c}{Images only} & \multicolumn{2}{c}{Images $+$ text}
      & \multicolumn{2}{c}{Distilled} & & \\
    \cmidrule(lr){5-6}\cmidrule(lr){7-8}\cmidrule(lr){9-10}
    Method & Anchor & $d$ & Stage 2
      & DINOv3 & Franca & CLIP & SigLIP2 & PEspatial & RADIO & Mean & $\Delta$ \\
    \midrule
    \multicolumn{12}{l}{\textit{$\mathcal{J}$-Mean $\uparrow$}} \\
    NAF~\cite{chambon2026naf} & -- & 256 & \checkmark & \underline{66.00} & \underline{68.70} & 62.80 & 62.20 & \underline{67.00} & \underline{66.40} & \underline{65.52} & -- \\
    NAF & -- & 128 & \checkmark & 65.60 & 68.40 & \underline{62.90} & 61.90 & 66.80 & 65.90 & 65.25 & \dneg{0.27} \\
    NAF & -- & 128 & & 62.79 & 66.59 & 61.77 & \underline{62.31} & 65.88 & 65.02 & 64.06 & \dneg{1.46} \\
    \textbf{AnchorUp} & bilinear & 128 & & \textbf{66.30} & \textbf{69.50} & \textbf{63.50} & \textbf{62.40} & \textbf{67.70} & \textbf{67.20} & \textbf{66.10} & \dpos{0.58} \\
    \midrule
    \multicolumn{12}{l}{\textit{$\mathcal{F}$-Mean $\uparrow$}} \\
    NAF~\cite{chambon2026naf} & -- & 256 & \checkmark & \underline{71.40} & 74.40 & 68.50 & \underline{68.30} & 73.60 & \underline{71.10} & \underline{71.22} & -- \\
    NAF & -- & 128 & \checkmark & 70.90 & \underline{74.70} & \underline{68.70} & 68.20 & \underline{73.70} & 70.60 & 71.13 & \dneg{0.09} \\
    NAF & -- & 128 & & 67.41 & 71.82 & 66.49 & 67.28 & 71.63 & 69.41 & 69.01 & \dneg{2.21} \\
    \textbf{AnchorUp} & bilinear & 128 & & \textbf{71.90} & \textbf{75.60} & \textbf{69.00} & \textbf{68.70} & \textbf{74.80} & \textbf{72.00} & \textbf{72.00} & \dpos{0.78} \\
    \midrule
    \multicolumn{12}{l}{\textit{$\mathcal{J}\&\mathcal{F}$ $\uparrow$}} \\
    NAF~\cite{chambon2026naf} & -- & 256 & \checkmark & \underline{68.70} & \underline{71.55} & 65.65 & \underline{65.25} & \underline{70.30} & \underline{68.75} & \underline{68.37} & -- \\
    NAF & -- & 128 & \checkmark & 68.25 & \underline{71.55} & \underline{65.80} & 65.05 & 70.25 & 68.25 & 68.19 & \dneg{0.18} \\
    NAF & -- & 128 & & 65.10 & 69.20 & 64.13 & 64.80 & 68.75 & 67.22 & 66.53 & \dneg{1.84} \\
    \textbf{AnchorUp} & bilinear & 128 & & \textbf{69.10} & \textbf{72.50} & \textbf{66.20} & \textbf{65.50} & \textbf{71.20} & \textbf{69.60} & \textbf{69.02} & \dpos{0.65} \\
    \bottomrule
  \end{tabular}
  \caption{\textbf{Semi-supervised video object segmentation on DAVIS-2017.}
  Label propagation uses the upsampled features directly, without any training on
  video. Rows follow the order of Table~\ref{tab:main}, each differing from the
  row above it in one factor, so the third NAF row differs from \method{} only by
  Eq.~\ref{eq:anchorup}. %
  The third NAF row is trained using Stage 1 only, following the \method{} recipe.
  The gain is largest on $\mathcal{F}$-Mean, the boundary-accuracy term
  (Section~\ref{sec:analysis} shows the correction concentrates on boundaries).
  Best per column in bold, second best underlined; $\Delta$ is the
  difference in mean from the two-stage NAF row.
  }
  \label{tab:davis}
\end{table*}

%% file: tables/tab_efficiency.tex
\begin{table}[t]
  \centering
  \small
  \setlength{\tabcolsep}{3pt}
  \resizebox{\linewidth}{!}{%
  \begin{tabular}{l r r r}
    \toprule
    Method & Params $\downarrow$ & Latency (ms) $\downarrow$ & Peak mem.\ (MiB) $\downarrow$ \\
    \midrule
    Bilinear & 0 & 0.30 & 297.4 \\
    Bicubic & 0 & 0.84 & 297.4 \\
    FeatUp~\cite{fu2024featup} & 173{,}540 & 27.95 & 2674.0 \\
    JAFAR~\cite{couairon2025jafar} & 628{,}480 & 45.82 & 5410.9 \\
    AnyUp~\cite{wimmer2025anyup} & 845{,}056 & 49.27 & 3253.8 \\
    NAF~\cite{chambon2026naf} & 662{,}528 & 25.31 & 1685.5 \\
    \textbf{AnchorUp} (bilinear) & \textbf{167{,}425} & \textbf{16.21} & \textbf{1585.2} \\
    \textbf{AnchorUp} (bicubic) & \textbf{167{,}425} & \textbf{16.71} & \textbf{1577.1} \\
    \bottomrule
  \end{tabular}}
  \caption{\textbf{Cost of one upsampling pass} for the upsampler alone, at
  $448{\times}448$ output from the $28{\times}28$ features of 384 channels that a
  $448{\times}448$ input produces, FP32 and batch 1 on an RTX PRO 6000 Blackwell.
  Latencies are means over 500 forwards and the standard deviation is below
  $0.07$\,ms for every row. %
  \method{} is $1.6\times$ faster than NAF, with a
  quarter of its parameters.}
  \label{tab:efficiency}
\end{table}

%% file: tables/tab_seeds.tex
\begin{table}[t]
  \centering
  \small
  \setlength{\tabcolsep}{4pt}
  \begin{tabular}{l ccc}
    \toprule
    Method & VOC & Cityscapes & DAVIS \\
    \midrule
    Control & $81.36 \pm 0.10$ & $59.44 \pm 0.70$ & $65.32 \pm 1.05$ \\
    \textbf{AnchorUp} & $81.19 \pm 0.18$ & $60.57 \pm 0.08$ & $68.87 \pm 0.15$ \\
    \midrule
    Difference & \dneg{$0.16 \pm 0.09$} & \dpos{$1.14 \pm 0.62$} & \dpos{$3.54 \pm 0.92$} \\
    \bottomrule
  \end{tabular}
  \caption[Comparison with the control over three training seeds]{\textbf{Comparison with the control over three training seeds.} The control is the third NAF row of Tables~\ref{tab:main} and~\ref{tab:davis} ($d = 128$, stage 1 only), and \method{} uses the bilinear anchor. Each entry is the mean and standard deviation over seeds 0--2 of the mean across backbones (mIoU on VOC and Cityscapes, $\mathcal{J}\&\mathcal{F}$ on DAVIS); the difference is taken between models of the same seed. Seed 0 is the model from Tables~\ref{tab:main} and~\ref{tab:davis}, with the linear probes retrained.}
  \label{tab:seeds}
\end{table}

%% file: sec/5_analysis.tex
\section{What the Learned Part Learns}
\label{sec:analysis}

Eq.~\ref{eq:anchorup} defines what the learned part is fitted to. Writing its
contribution as $C_\theta = \alpha(\Flearn - \Fanchor)$, the output is
$\Fhat = \Fanchor + C_\theta$, so the reconstruction loss is
\begin{equation}
  \bigl\lVert \Fhat - \Fhr \bigr\rVert^2
  = \bigl\lVert C_\theta - (\Fhr - \Fanchor) \bigr\rVert^2 .
  \label{eq:residual}
\end{equation}
The term in parentheses is the residual the anchor leaves: the learned part is
fitted to it, not to the feature map. Where the anchor is already right, the residual is small, so the loss requires
only a small contribution. %
Whether this residual is high-frequency and localized near boundaries depends on the target and the anchor,
not on Eq.~\ref{eq:residual}. We measure this property on DINOv3 features by
decomposing the trained model's prediction into $\Fanchor$, $\Flearn$, and $C_\theta$.

Figure~\ref{fig:spectrum} plots the Fourier power of each term of
Eq.~\ref{eq:anchorup}. Every term is produced at image resolution, one feature
per pixel, so the frequency axis is in cycles per pixel. At the lowest
measured frequency the correction carries over two orders of magnitude less power
than the anchor (about $10^6$ against $2 \times 10^8$): the low-frequency content of the output comes almost entirely
from the anchor.
The correction carries more power than the anchor between roughly $0.06$ and
$0.3$ cycles per pixel.

Figure~\ref{fig:pca} shows the same decomposition in image space. The correction is flat
over interiors and background and structured where the anchor is blurred: contours,
thin repeated structure, the boundary between an object and its shadow. Band-passing the correction makes this concrete (see supplementary material): the 0.06–0.3 band, where the correction carries more power than the anchor, holds the fine contour lines; the frequencies below hold the broad response around the same boundaries; the frequencies above hold almost nothing.

These two properties settle what Section~\ref{sec:method-anchorup} could only
hypothesize: the correction is a far smaller target than the feature map. A
model that predicts $\Fhr$ directly must produce the smooth content itself.
\method{} leaves it to the anchor and saves that capacity.
Because what remains is boundary work, the clearest gain falls where boundaries decide
the score: the contour accuracy of DAVIS (Section~\ref{sec:davis}).

%% file: sec/6_conclusion.tex
\section{Limitations}
\label{sec:limitations}

Each term of Eq.~\ref{eq:anchorup} admits a broader choice than we have tested. The
anchor is one of two classical interpolants, even though
Section~\ref{sec:crossbackbone} shows the better one is benchmark-dependent. The gate
is one scalar, uniform even where the anchor's reliability varies, for example,
between flat regions and object boundaries. The learned part is NAF's upsampler, yet
the framework accepts any learned upsampler in its place. Searching over these
choices is left to future work.

Our evaluation also stays within natural-image benchmarks. Domains that separate
cleanly into flat and structured regions, such as remote sensing, are where we
expect the anchoring premise to be strongest, yet we have not tested them.

\section{Conclusion}
\label{sec:conclusion}

We have presented \method{}, a learned feature upsampler
that predicts a correction to a classical interpolant rather than the high-resolution
feature map itself. The trained correction is what this framing asks for: high-frequency,
and concentrated at the boundaries where interpolation fails. Thus, with a quarter of
NAF's parameters and only the first of its two training stages, \method{} matches NAF
on VOC and Cityscapes, outperforms it on DAVIS-2017 across all backbones, and runs
1.6$\times$ faster. The result argues for the framing more
than for the architecture. When a parameter-free interpolation already covers most of the
target, the model should predict what is missing, not what is already there.

%% file: sec/7_supp.tex
The main paper defers full derivations, experimental details, and supporting
measurements to this supplementary material. We provide them here in the order they
are referenced in the main paper.
Sections~\ref{sec:supp-algorithm}--\ref{sec:supp-bound} give the method in
full: the forward pass, the trained gate against its closed form, and  how the learned part's error reaches the output.
Sections~\ref{sec:supp-backbones}--\ref{sec:supp-training} detail the
experimental setup: the frozen backbone checkpoints, the evaluation protocols,
and the training settings. Section~\ref{sec:supp-seeds} gives the per-backbone
values behind Table~5 of the main paper.
Sections~\ref{sec:supp-anchor-backbone}--\ref{sec:supp-steps} extend the
results: which backbones gain from the bicubic anchor, how accuracy scales
with the guidance dimension, and what longer training would buy.
Section~\ref{sec:supp-spectrum-protocol} covers the spectral protocols behind
Figures~3 and~4 of the main paper.
Section~\ref{sec:supp-qualitative} shows the qualitative results at full
length.

\section{Forward Pass}
\label{sec:supp-algorithm}

Section~3.2 of the main paper defers the forward pass to this section.
Algorithm~\ref{alg:anchorup} lists it in full: the anchor, the NAF-style
prediction, and the gated combination of Eq.~1 of the main paper. We
write $g_\theta$ for the guidance network of Section~3.3 of the main paper, and
$\mathrm{NA}_k$ for neighborhood cross-attention over a $k \times k$ window.

\begin{algorithm}[t]
\caption{\method{} forward pass}
\label{alg:anchorup}
\begin{algorithmic}[1]
\Require image $I$, features $\Flr$, output size $(H,W)$, guidance network
  $g_\theta$, gate $\gamma$
\State $\Fanchor \gets u(\Flr,\,(H,W))$
\State $Q \gets \mathrm{RoPE}\bigl(g_\theta(I)\bigr)$
\State $K \gets \mathrm{avgpool}_{h\times w}(Q)$,\quad $V \gets \Flr$
\State $\Flearn \gets \mathrm{NA}_{k}(Q, K, V)$
\State $\alpha \gets \sigma(\gamma)$
\State \Return $\Fanchor + \alpha\,(\Flearn - \Fanchor)$
\end{algorithmic}
\end{algorithm}

\section{Closed-Form Gate in Full}
\label{sec:supp-alpha}

\input{tables/tab_alpha}

\begin{figure}[t]
  \centering
  \includegraphics[width=\linewidth]{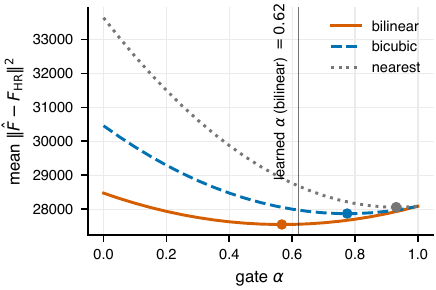}
  \caption{\textbf{The error is flat around the learned gate $\alpha$, and
  its minimum moves with the anchor.} The loss of Section~3.4 of the main
  paper as a function of $\alpha$, averaged over 200 Cityscapes validation
  images, with the guidance network of the bilinear-anchor model frozen
  and the anchor swapped. Dots mark minima; the vertical line marks the
  learned gate. The norm sums over all feature elements of an image,
  setting the scale of the axis.}
  \label{fig:supp-alpha}
\end{figure}

This section examines the gate $\alpha$ from three directions: whether
training finds the right value, how much the exact value matters, and how
the optimum moves with the anchor. First, does
the trained $\alpha$ land near the optimum $\alpha^\star$ that Eq.~3 of the
main paper derives? Table~\ref{tab:alpha} reports the trained $\alpha$ and
the optimum $\alpha^\star$ together. On \method{}'s training data, the two agree by construction: because the
loss is quadratic, a converged gate lies exactly at $\alpha^\star$
(Section~3.4 of the main paper). We therefore solve Eq.~3 on 200 held-out Cityscapes
images, where nothing forces agreement.

With both anchors, the trained
$\alpha$ lands just above $\alpha^\star$: $0.620$
against $0.551$ for bilinear, $0.771$ against $0.748$ for bicubic.
The ordering tracks how much of the answer each anchor already holds: bicubic
reconstructs the target less well, so its $\alpha^\star$ lies further from the
anchor, and the trained $\alpha$ follows. Whether the remaining offsets matter depends
on how sharply the error rises away from $\alpha^\star$, which
Figure~\ref{fig:supp-alpha} measures next.

Second, how sensitive is the reconstruction error to the exact value of
$\alpha$? The solid bilinear curve in Figure~\ref{fig:supp-alpha} steps
$\alpha$ from 0 to 1 in increments of 0.05
and traces the error of every mixture the gate could have
chosen; at $\alpha = 0$ the curve starts at the anchor's own error. The
curve is shallow: across $\alpha \in [0.4,\,0.8]$, the
error varies by $0.56\%$, and both the trained $\alpha$ and the optimum
$\alpha^\star$ fall well inside this flat stretch. Offsets of the size shown
in Table~\ref{tab:alpha} are therefore inconsequential; the gate is not a
sensitive hyperparameter. Consistent with this, the main paper never tunes
$\alpha$ per benchmark: every model runs with the value determined during its
training.

Third, how does $\alpha^\star$ move as the anchor changes? Answering this
needs the anchor to be the only thing that changes, so the other
two curves keep the guidance network $g_\theta$ of the bilinear-anchor model
frozen while swapping the anchor
underneath it: bicubic, and nearest neighbor as a deliberately weaker anchor
that extends the quality axis downward. No gate was trained for these
anchors, which is why only the bilinear curve carries a learned-gate mark.
Two patterns emerge. The curves keep their order at every $\alpha$. That
order is set at $\alpha = 0$, where the mixture reduces to the anchor alone
and the intercept is each anchor's own error. And the minimum of the
nearest-neighbor curve lies at $\alpha^\star = 0.93$ (the dot): as the
anchor worsens, $\alpha^\star$ moves toward the learned prediction, until a bad
anchor is discarded rather than corrected.

\section{\texorpdfstring{How the Learned Part's Error Reaches the Output}{How the Learned Part's Error Reaches the Output}}
\label{sec:supp-bound}

 Section~3.5 of the main paper states that, in \method{}, an error in the learned part at a resolution unseen during training enters the output scaled by $\alpha$. This section derives the bound behind that statement. Subtracting $\Fhr$ from both sides of Eq.~1 of the main paper gives
\begin{equation}
  \Fhat - \Fhr
  = (1-\alpha)\bigl(\Fanchor - \Fhr\bigr) + \alpha\bigl(\Flearn - \Fhr\bigr),
\end{equation}
so by the triangle inequality
\begin{equation}
  \lVert \Fhat - \Fhr \rVert
  \le (1-\alpha)\,\lVert \Fanchor - \Fhr \rVert
  + \alpha\,\lVert \Flearn - \Fhr \rVert.
  \label{eq:supp-bound}
\end{equation}
The output error is a convex mixture of the two errors. NAF's second training stage
exists to keep $\Flearn$ well behaved when the upsampling factor at evaluation
differs from the $\times 2$ seen in training.  In \method{}, the anchor term is computed by the same interpolation
rule at every factor, so when an unseen factor degrades $u_\theta$, at most the
$\alpha$-weighted share of the output degrades with it.

\section{Backbones}
\label{sec:supp-backbones}

The tables of the main paper name each backbone only by family.
Table~\ref{tab:supp-backbones} names the exact checkpoint behind each column
heading. NAF and
\method{} are trained once on DINOv3-B/16 and applied unchanged to the other
backbones; AnyUp is trained once on DINOv2 ViT-S, the feature extractor of its
official configuration, and applied unchanged; FeatUp and JAFAR are trained per
backbone, as their official configurations require.

\begin{table}[t]
  \centering
  \scriptsize
  \setlength{\tabcolsep}{3pt}
  \begin{tabular}{l l c}
    \toprule
    Backbone & Checkpoint & Patch \\
    \midrule
    DINOv3    & \texttt{vit\_base\_patch16\_dinov3.lvd1689m}    & 16 \\
    CLIP      &  & 16 \\
              & \texttt{vit\_base\_patch16\_clip\_384.}  &    \\
              & \texttt{\ \ laion2b\_ft\_in12k\_in1k}    &    \\
    SigLIP2   & \texttt{vit\_base\_patch16\_siglip\_512.v2\_webli} & 16 \\
    Franca    & \texttt{franca\_vitb14}                         & 14 \\
    PEcore    & \texttt{vit\_pe\_core\_base\_patch16\_224.fb}   & 16 \\
    PEspatial & \texttt{vit\_pe\_spatial\_base\_patch16\_512.fb} & 16 \\
    RADIO     & \texttt{radio\_v2.5-b}                          & 16 \\
    \bottomrule
  \end{tabular}
  \caption{\textbf{Frozen backbones.} All are base-size vision transformers; Franca uses
  a patch size of 14 and the rest 16.}
  \label{tab:supp-backbones}
\end{table}

The main paper groups these seven by their pretraining signal: image-only,
image--text, and distilled. One assignment needs a note. The base-size PEcore
checkpoint is itself finetuned by distilling image--text similarity
distributions from PEcoreG; we group it by what that signal carries, which is
image--text, and keep the distilled group for the two models whose teachers
supply spatial structure.

\section{Evaluation Protocols}
\label{sec:supp-protocol}

Section~4 of the main paper describes each evaluation in a sentence and
defers the full protocols to this section: the linear probe, the DAVIS label
propagation, and the timing methodology, together with the anchor-cost
comparison the timing supports.

\paragraph{Semantic segmentation.}
We freeze the backbone and the upsampler and train a single linear layer on the
upsampled features:  20 epochs, AdamW at $5\times10^{-4}$ with a cosine schedule, batch size 2, random horizontal flips, images at
$448\times448$, mIoU on the validation split. We add no auxiliary loss. The probe is deliberately the weakest readout
that can expose a difference in the features.

\paragraph{Video object segmentation.}
On DAVIS-2017 we propagate the first-frame masks through the upsampled features
directly, with no training on video. Each frame matches against the first frame
and the previous 7, within a spatial neighborhood of 12 pixels, and takes the top
5 matches. The upsampling factor is $\times 2$. We report $\mathcal{J}$-Mean,
$\mathcal{F}$-Mean and their mean over the six backbones of Table~3 of the main
paper, which exclude PEcore.

\paragraph{Cost.}
We measure the upsampler alone, with the backbone excluded, so these are not
end-to-end image processing times. Every configuration in Table~4 of the main
paper comes from one session on one GPU with the same input tensors: an NVIDIA
RTX PRO 6000 Blackwell Max-Q Workstation Edition, PyTorch 2.6.0 with CUDA 12.8,
FP32, batch size 1, under \texttt{torch.inference\_mode()}. The RGB input is
$1{\times}3{\times}448{\times}448$, the feature input
$1{\times}384{\times}28{\times}28$ and the output
$1{\times}384{\times}448{\times}448$. We run 20 warm-up forwards, then time 100
forwards five times over, and report the mean and population standard deviation of
those 500. CUDA events bracket each forward, and we synchronize before reading the
elapsed time; we do not empty the cache between timed forwards. Peak memory is
\texttt{torch.cuda.max\_memory\_allocated()} over the model, inputs, outputs and
temporaries, which is not the memory the driver reserves. We do not load
checkpoints, since weight values do not change the shape of the execution graph.

These numbers hold for that GPU, dtype, batch size and output resolution, and do
not carry over to others. The standard deviation describes variation across the
500 forwards, not across machines or processes.

\paragraph{Cost of the anchor.}
Table~4 of the main paper carries the comparison: switching \method{} from
bilinear to bicubic adds $0.51$\,ms, while bilinear and bicubic run on their
own differ by $0.54$\,ms (both from the unrounded session measurements). The
two agree to within $0.04$\,ms, so the whole cost of the anchor change is the
interpolation itself: the attention path and the parameter count are
untouched, and the peak memory moves by about $0.5$\%. Even with the more
expensive anchor, \method{} stays $34.0$\% below NAF in latency and $6.4$\%
below in peak memory. Which anchor is preferable is therefore an accuracy
question; Section~4.1 of the main paper settles it, not the timing.

\section{Training Configuration}
\label{sec:supp-training}

Section~3.5 of the main paper summarizes the training recipe;
Table~\ref{tab:supp-training} gives the full setup of the models the main
paper reports. The two anchors share every setting. The model is the only thing
that differs from the configuration distributed with NAF's code; every optimizer, data and schedule
setting is that code's default.

One of those defaults departs from the NAF paper: the paper reports a batch
size of 2 for this stage (its Appendix~B.1), while the configuration shipped
with the released code uses 4. We follow the code. Every model we train ---
including the retrained NAF rows of Tables~1 and~3 of the main paper --- uses
the same batch size, so no comparison mixes the two.

One check that the retrained NAF rows are faithful: at $d = 128$, NAF's second
stage adds $0.49$ mIoU on VOC with DINOv3 (the two-stage row of Table~1 of the
main paper against the stage-one row), in line with the $+0.4$ its authors
report at $d = 256$ (their Appendix~B.1). The published effect reproduces in
our environment.

\begin{table}[t]
  \centering
  \small
  \setlength{\tabcolsep}{4pt}
  \begin{tabular}{l l}
    \toprule
    Setting & Value \\
    \midrule
    Data                & ImageNet, $512\times512$ HR targets \\
    Degradation         & fixed, $\times 2$ \\
    Guidance resolution & $\min(224,\, 4\times\text{HR feature})$ per axis \\
    Loss                & $\ell_2$ feature reconstruction, no auxiliary terms \\
    Batch size          & 4 \\
    Optimizer           & AdamW, $\beta=(0.9,\,0.999)$ \\
    Learning rate       & $2\times10^{-4}$ \\
    Weight decay        & $10^{-5}$ \\
    Seed                & 0 (also 1 and 2 in Section~\ref{sec:supp-seeds}) \\
    Reported checkpoint & 25{,}000 steps \\
    \midrule
    Anchor              & bilinear or bicubic, one model each \\
    Guidance dimension $d$  & 128 \\
    Attention heads     & 4 \\
    RoPE heads          & 4 \\
    RoPE base / rescale & 100 / 2 \\
    Neighborhood $k$   & 9 \\
    Encoder layers      & 2 \\
    Gate                & $\sigma(\gamma)$, $\gamma$ initialized at 0 \\
    \bottomrule
  \end{tabular}
  \caption{\textbf{Training setup in full.} The position encoding is worth
  stating explicitly because a later default in the same codebase disables the
  RoPE rescale: these results use rescale 2.}
  \label{tab:supp-training}
\end{table}

\section{Training Seeds}
\label{sec:supp-seeds}

Sections~4.1 and~4.2 of the main paper state that the differences between
\method{} (bilinear) and the control keep their signs over three training seeds
(Table~5 of the main paper); Table~\ref{tab:supp-seeds} gives the values for each
backbone. Seed 0 is the model from Tables~1 and~3 of the main paper. The linear probes
in this comparison are retrained, so the seed-0 means over the seven backbones
differ from Table~1 by at most 0.11 mIoU. Label propagation involves no training, and the
seed-0 DAVIS scores match Table~3 up to rounding.

\input{tables/tab_supp_seeds}

\section{The Anchor-Kernel Gain across Backbones}
\label{sec:supp-anchor-backbone}

Section~4.1 of the main paper reports that the bicubic anchor helps some backbones
and hurts others, and that what separates the two groups is how a backbone was
pretrained --- with or without image--text supervision --- rather than how
accurate it is. This section shows the per-backbone numbers behind that claim
and how far they support it.

Table~\ref{tab:supp-anchor} gives the gain for
every backbone, taken from
Table~1 of the main paper as the bicubic row minus the bilinear row. The split is
clean: the three image--text backbones take the top three places on both
benchmarks, and the four others take the bottom four, with no overlap.

\begin{table}[h]
  \centering
  \small
  \setlength{\tabcolsep}{5pt}
  \begin{tabular}{l c r r}
    \toprule
    Backbone & Image--text & $\Delta$ VOC & $\Delta$ Cityscapes \\
    \midrule
    PEcore    & \checkmark & $+1.98$ & $+0.42$ \\
    SigLIP~2  & \checkmark & $+1.73$ & $+0.30$ \\
    CLIP      & \checkmark & $+0.55$ & $-0.29$ \\
    \midrule
    RADIO     &            & $+0.03$ & $-1.12$ \\
    PEspatial &            & $-0.03$ & $-0.79$ \\
    Franca    &            & $-0.07$ & $-0.51$ \\
    DINOv3    &            & $-0.41$ & $-0.92$ \\
    \bottomrule
  \end{tabular}
  \caption{\textbf{Bicubic anchor minus bilinear anchor, by backbone.} A check
  under \emph{Image--text} marks the backbones pretrained with image--text
  supervision. Rows are
  ordered by the VOC gain. The three image--text backbones take the
  top three places on both benchmarks and the four others take the bottom
  four, with no overlap. One caveat for the PEcore--PEspatial reading of
  Section~4.1 of the main paper: the two checkpoints differ in nominal
  resolution as well, 224 and 512.}
  \label{tab:supp-anchor}
\end{table}

Two limits temper this reading: the grouping is drawn after the fact over
seven backbones, and the explanation does not predict the order inside the
top group, where CLIP and PEcore, the two backbones with the most similar
pretraining, land at opposite ends. The split itself is what the table shows on both
benchmarks; the pretraining account is the interpretation.

\section{Guidance Dimension and Anchor}
\label{sec:supp-ablation}

\input{tables/tab_ablation}

Table~\ref{tab:ablation} provides the full sweep that Section~4.3 of the main
paper summarizes: \method{} trained at four guidance dimensions for both
anchors, with the training recipe fixed. From $d = 32$ to $128$, the
four-backbone mean rises by
about one point ($58.23 \to 59.26$ bilinear, $58.28 \to 59.14$ bicubic);
doubling again to $256$ buys an additional $0.10$ and $0.20$, at four times
the parameter count. The main paper reports $d = 128$, the knee of this
trade; at that size, \method{} also runs at a quarter of the published NAF's
parameters.

The $\alpha$ column adds a second reading: across an
$8\times$ capacity range, the trained gate moves by at most about $0.05$,
showing that capacity changes accuracy, not the division of labor between the
anchor and the prediction. The $d = 128$ gates are the ones compared against
their closed forms in Section~\ref{sec:supp-alpha}.

\section{Training past the Reported Checkpoint}
\label{sec:supp-steps}

\input{tables/tab_steps}

\begin{figure}[t]
  \centering
  \includegraphics[width=\linewidth]{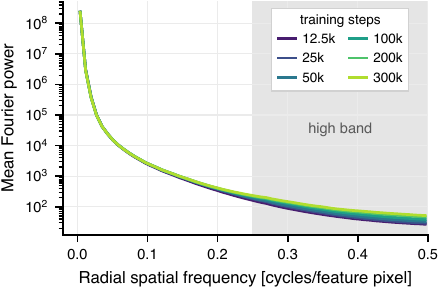}
  \caption{\textbf{Where the extra training goes.} Radially averaged power of the
  \method{} prediction with a bilinear anchor, at six checkpoints of the same run.
  The band below $0.05$ cycles per feature pixel is settled at the first
  checkpoint and moves by less than $0.5\%$ thereafter; all of the growth is in
  the shaded band above $0.25$, which gains $29\%$ between 25{,}000 and
  100{,}000 steps and $56\%$ between 25{,}000 and 300{,}000.}
  \label{fig:supp-steps}
\end{figure}

The main paper stops training at 25{,}000 steps. What would additional
training buy, and why stop there? This section runs the same configuration to
300{,}000 steps: the high-frequency band of the prediction keeps growing
(Figure~\ref{fig:supp-steps}), and the growth registers where boundaries
decide the score (Table~\ref{tab:supp-steps}).

Figure~\ref{fig:supp-steps} tracks the spectrum over checkpoints. The low-frequency band is fixed by the
anchor and does not move, which is the mechanism of Eq.~2 seen over training time
rather than over a single forward pass. The high band grows steadily and has not
leveled off at 300{,}000 steps.

Table~\ref{tab:supp-steps} shows where the growth lands. Four times the
training moves VOC by $+0.02$ mIoU, Cityscapes by $+0.17$ and DAVIS by
$+0.41$. The gains are
ordered by how much each benchmark depends on boundaries. Cityscapes is
dominated by thin structures (Section~4.1 of the main paper), so the extra
high-frequency content (sharper edges) pays at its boundaries. DAVIS
propagates labels through feature matches at boundaries (Section~4.2 of the
main paper), so it rewards the same growth more. VOC's large objects leave
little for sharper boundaries to win. The reported model, and every
comparison in the paper, remains the 25{,}000-step checkpoint. Stopping there
keeps the training budget equal to NAF's first stage --- at twelve times the
compute, every baseline would also move --- and the longer runs serve only to
show where additional training goes.

\section{Protocols for the Spectral Analyses of the Main Paper}
\label{sec:supp-spectrum-protocol}

\begin{figure}[t]
  \centering
  \begin{minipage}{\linewidth}
    \centering
    \newcommand{\bandhead}[1]{\makebox[\dimexpr(\linewidth-8pt)/3\relax][c]{\footnotesize #1}}%
    \bandhead{Below $0.065$}\bandhead{$0.065$--$0.289$}\bandhead{Above $0.289$}\\[1.5pt]
    \includegraphics[width=\linewidth]{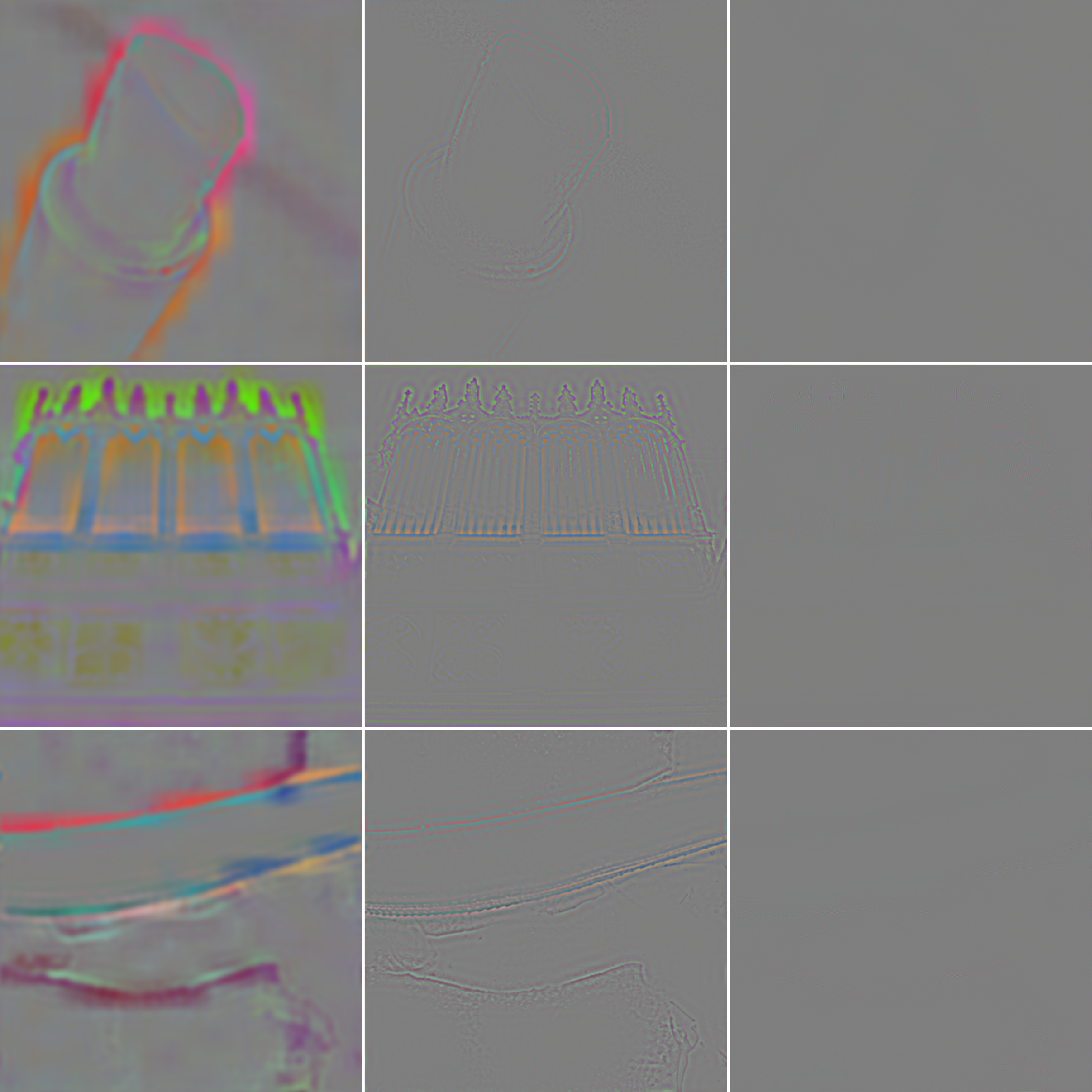}
  \end{minipage}
  \caption{\textbf{The correction split at the crossings of Figure~4 of the
  main paper.} Rows are the three scenes of its Figure~3. The middle band, where the
  correction carries more power than the anchor, holds the fine contour lines;
  the band below holds the broad response around the same boundaries; the band
  above holds almost nothing.}
  \label{fig:supp-bands}
\end{figure}

This section details how the two spectral readings in Section~5 of the main
paper were produced: the decomposition in Figure~4, and the band split of
the correction behind the $0.06$--$0.3$ statement
(Figure~\ref{fig:supp-bands}).

\paragraph{Decomposition of Figure~4.}
Figure~4 of the main paper decomposes the reported model --- bilinear anchor,
$d = 128$, 25{,}000 steps --- on 100 ImageNet images drawn with seed 0, with every
term of Eq.~1 produced at $512\times512$. We average the two-dimensional Fourier
power $\lvert\hat F\rvert^2$ of each channel over the 768 channels, the 100
images, and the radial direction, in 64 uniform bins over $0$--$0.5$ cycles
per pixel. The
axis is therefore in cycles per image pixel, not the coarser cycles per
feature pixel of Figure~\ref{fig:supp-steps}. We apply no window, no mean removal, and no normalization; the vertical
axis is raw mean power on a logarithmic scale.

\paragraph{The correction split by band.}
Figure~\ref{fig:supp-bands} band-passes the correction of Figure~3 of the main
paper at the two frequencies where its power crosses the anchor's: $0.065$ and
$0.289$ cycles per pixel (the main paper rounds these to $0.06$ and $0.3$). The
masks are complementary radial raised-cosine windows applied to the raw
feature map before the PCA projection, so the three bands sum to the
correction; all three are rendered in the PCA basis and scale of Figure~3.
The split is what Section~5 of the main paper reads off it: the crossing band
holds the fine contour lines; the band below, the broad response around the
same boundaries; the band above, almost nothing.

\section{Additional Qualitative Results}
\label{sec:supp-qualitative}

\begin{figure*}[p]
  \centering
  \begin{minipage}{0.70\textwidth}
    \centering
    \newcommand{\pcahead}[1]{\makebox[\dimexpr(\linewidth-2pt)/6\relax][c]{\footnotesize #1}}%
    \pcahead{Image}\pcahead{LR features}\pcahead{Anchor}\pcahead{Learned}%
    \pcahead{Correction}\pcahead{\method{}}\\[1.5pt]
    \includegraphics[width=\linewidth]{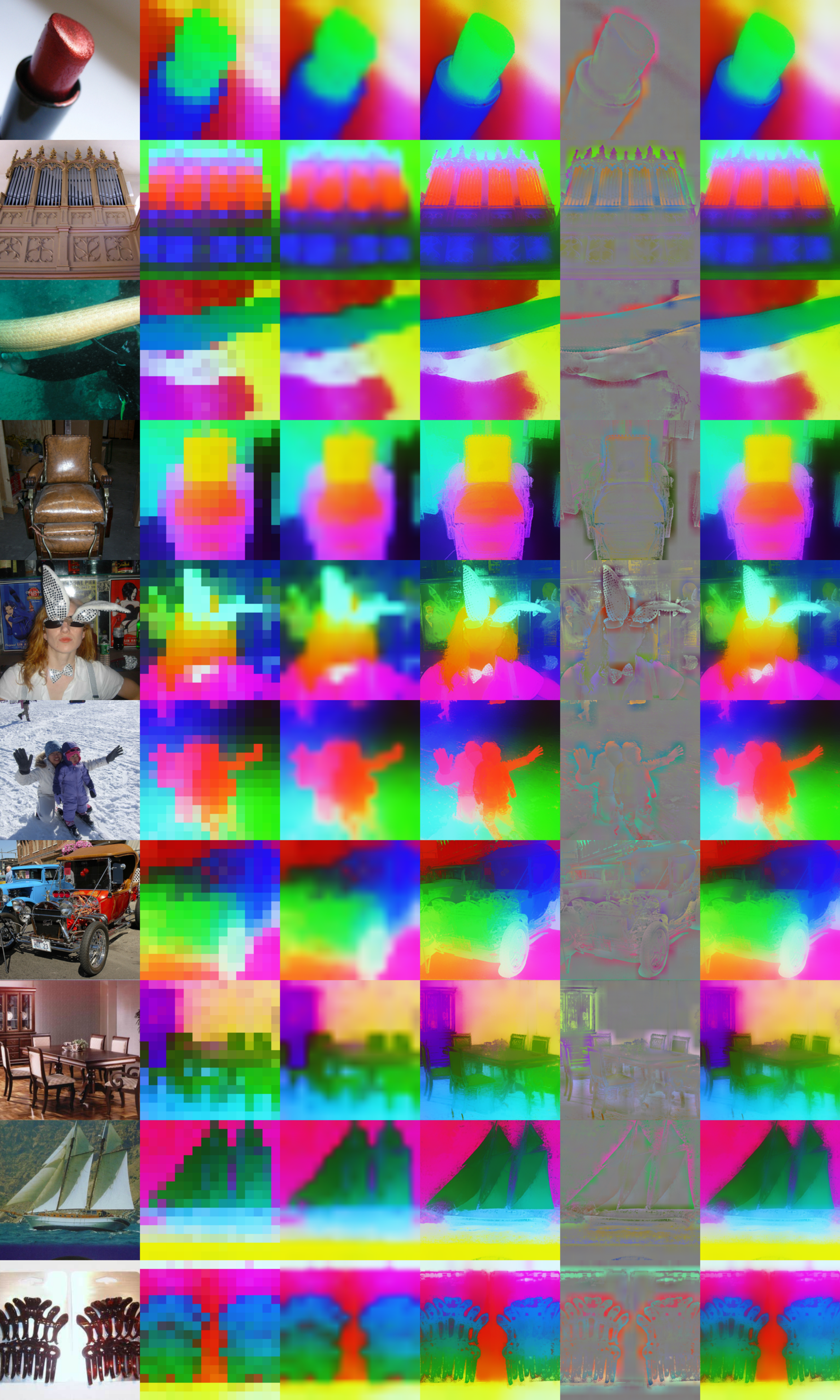}
  \end{minipage}
  \caption{\textbf{Decomposition, all ten images.} Columns as in Figure~3 of the
  main paper, which shows the first three rows.}
  \label{fig:supp-pca}
\end{figure*}

\begin{figure*}[p]
  \centering
  \begin{minipage}{0.70\textwidth}
    \centering
    \newcommand{\pcahead}[1]{\makebox[\dimexpr(\linewidth-2pt)/6\relax][c]{\footnotesize #1}}%
    \pcahead{Image}\pcahead{LR features}\pcahead{Anchor}\pcahead{Learned}%
    \pcahead{Correction}\pcahead{\method{}}\\[1.5pt]
    \includegraphics[width=\linewidth]{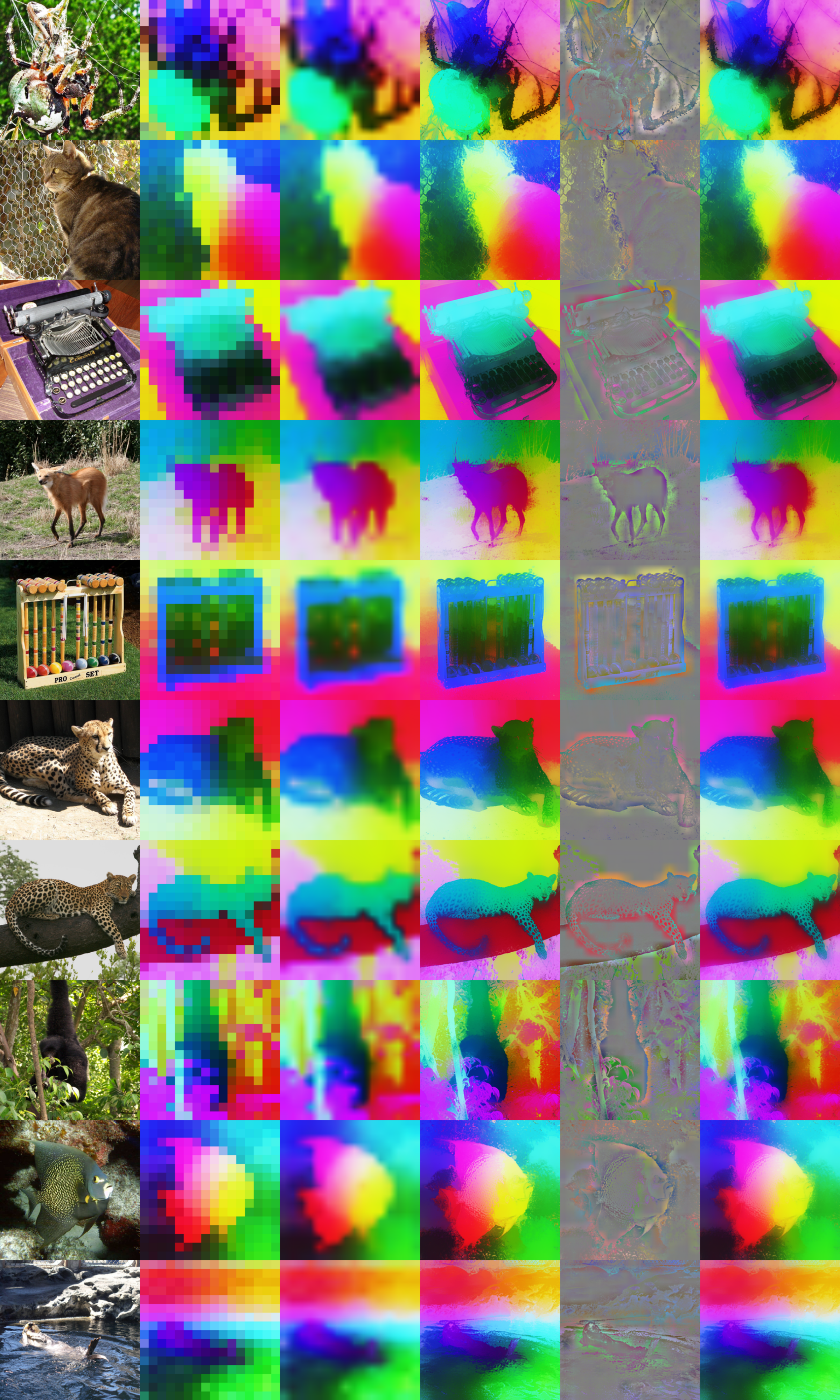}
  \end{minipage}
  \caption{\textbf{Decomposition on held-out images, rows 1--10 of 30.}
  ImageNet validation images, unseen during training; columns and projection
  as in Figure~\ref{fig:supp-pca}.}
  \label{fig:supp-pca-val-a}
\end{figure*}

\begin{figure*}[p]
  \centering
  \begin{minipage}{0.70\textwidth}
    \centering
    \newcommand{\pcahead}[1]{\makebox[\dimexpr(\linewidth-2pt)/6\relax][c]{\footnotesize #1}}%
    \pcahead{Image}\pcahead{LR features}\pcahead{Anchor}\pcahead{Learned}%
    \pcahead{Correction}\pcahead{\method{}}\\[1.5pt]
    \includegraphics[width=\linewidth]{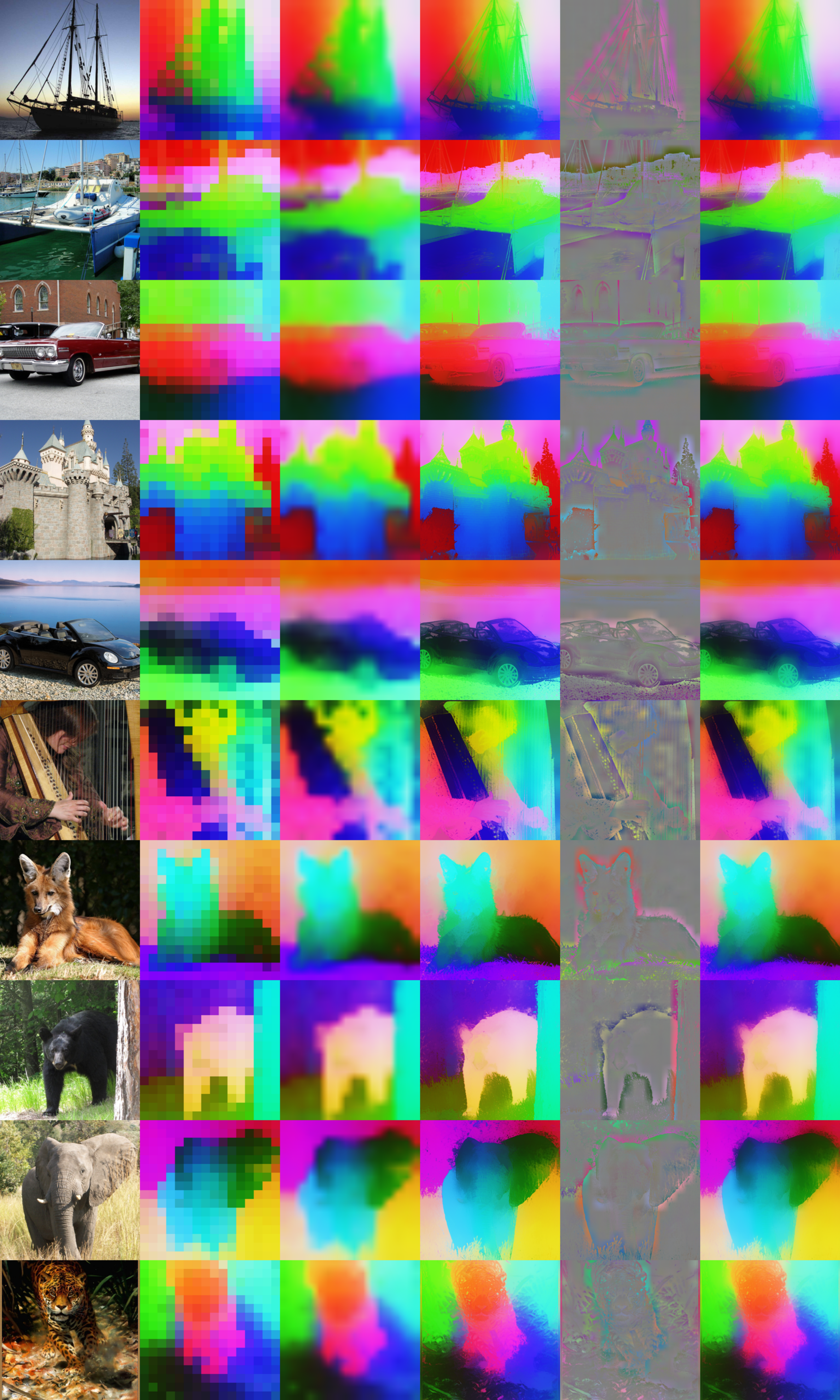}
  \end{minipage}
  \caption{\textbf{Decomposition on held-out images, rows 11--20 of 30.}
  Columns and projection as in Figure~\ref{fig:supp-pca}.}
  \label{fig:supp-pca-val-b}
\end{figure*}

\begin{figure*}[p]
  \centering
  \begin{minipage}{0.70\textwidth}
    \centering
    \newcommand{\pcahead}[1]{\makebox[\dimexpr(\linewidth-2pt)/6\relax][c]{\footnotesize #1}}%
    \pcahead{Image}\pcahead{LR features}\pcahead{Anchor}\pcahead{Learned}%
    \pcahead{Correction}\pcahead{\method{}}\\[1.5pt]
    \includegraphics[width=\linewidth]{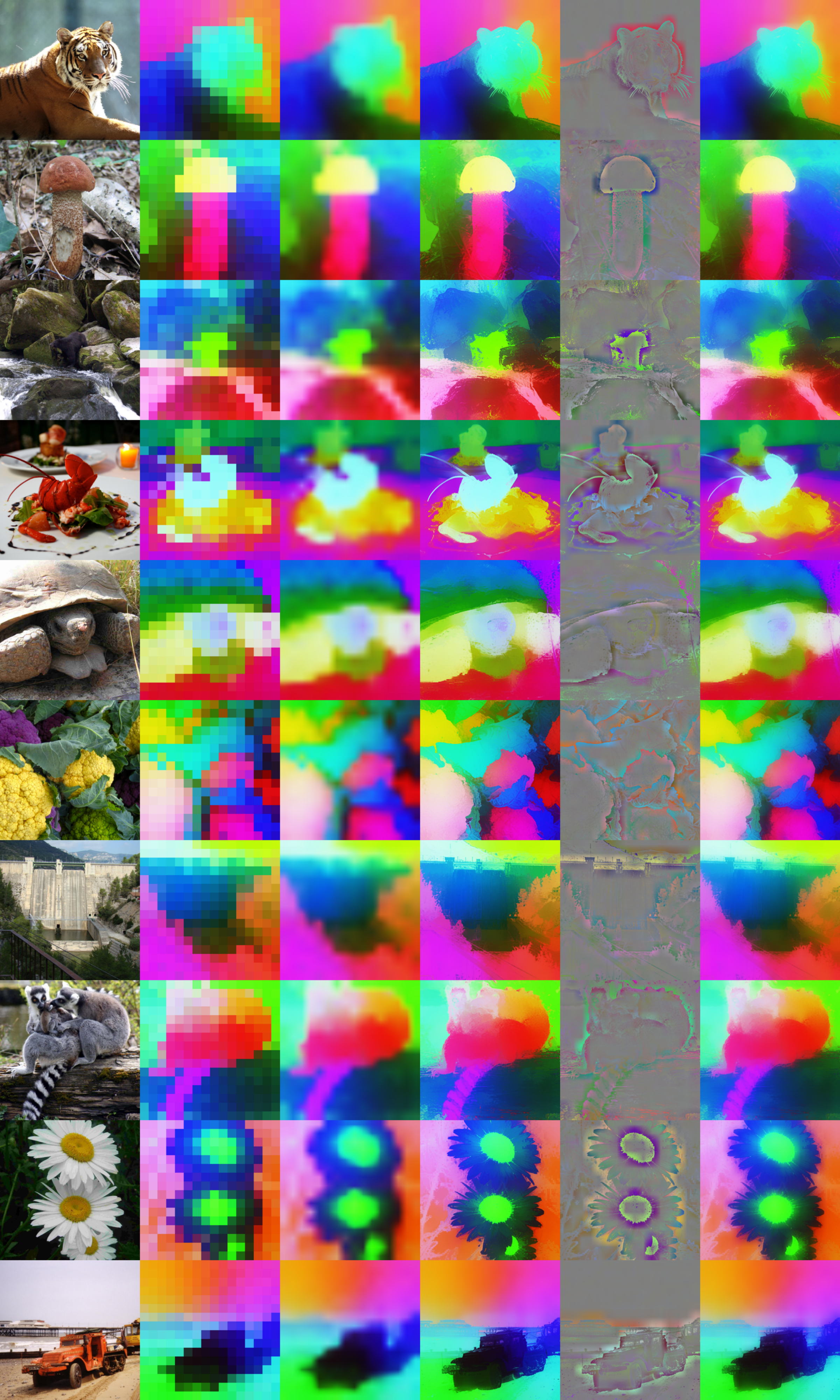}
  \end{minipage}
  \caption{\textbf{Decomposition on held-out images, rows 21--30 of 30.}
  Columns and projection as in Figure~\ref{fig:supp-pca}.}
  \label{fig:supp-pca-val-c}
\end{figure*}

Figure~3 of the main paper shows three scenes of its decomposition;
Figure~\ref{fig:supp-pca} shows all ten. One principal-component basis is
fitted per image and shared across the
columns, with the correction --- a zero-origin residual --- projected on the same
axes without mean subtraction. The ten rows extend the reading of Figure~3 of
the main paper without exception: in every scene the correction stays near
mid-gray over interiors and background, and its strength follows the amount of
fine structure.
Figures~\ref{fig:supp-pca-val-a}--\ref{fig:supp-pca-val-c} repeat the check on
30 ImageNet validation images, held out from training, with the same
projection; the reading does not change.

%% file: tables/tab_alpha.tex
\begin{table}[t]
  \centering
  \small
  \begin{tabular}{l c c c}
    \toprule
    Anchor & learned $\alpha$ & $\alpha^\star$ & difference \\
    \midrule
    Bilinear & 0.620 & 0.551 & $+0.069$ \\
    Bicubic & 0.771 & 0.748 & $+0.023$ \\
    \bottomrule
  \end{tabular}
  \caption{\textbf{The learned gate against the gate the loss would choose.}
  $\alpha^\star$ is Eq.~3 of the main paper solved on 200 Cityscapes validation
  images with DINOv3 features, no labels used, each row using its own trained
  model. The gate is trained by the same loss as the network, with no schedule
  and no separate term.}
  \label{tab:alpha}
\end{table}

%% file: tables/tab_supp_seeds.tex
\begin{table*}[t]
  \centering
  \footnotesize
  \setlength{\tabcolsep}{3pt}
  \begin{tabular}{l ccccccc c}
    \toprule
    & \multicolumn{2}{c}{Images only} & \multicolumn{3}{c}{Images + text} & \multicolumn{2}{c}{Distilled} & \\
    \cmidrule(lr){2-3} \cmidrule(lr){4-6} \cmidrule(lr){7-8}
    Method & DINOv3 & Franca & CLIP & SigLIP2 & PEcore & PEspatial & RADIO & Mean \\
    \midrule
    \multicolumn{9}{l}{\emph{(a) Pascal VOC}} \\
    Control & $86.93 \pm 0.35$ & $84.02 \pm 0.19$ & $80.70 \pm 0.15$ & $79.50 \pm 0.57$ & $74.50 \pm 0.76$ & $77.42 \pm 0.16$ & $86.45 \pm 0.09$ & $81.36 \pm 0.10$ \\
    \textbf{AnchorUp} & $87.58 \pm 0.13$ & $84.32 \pm 0.02$ & $80.47 \pm 0.20$ & $78.38 \pm 0.56$ & $73.26 \pm 0.64$ & $77.68 \pm 0.02$ & $86.67 \pm 0.02$ & $81.19 \pm 0.18$ \\
    Difference & \dpos{$0.65 \pm 0.22$} & \dpos{$0.30 \pm 0.18$} & \dneg{$0.23 \pm 0.06$} & \dneg{$1.12 \pm 0.01$} & \dneg{$1.24 \pm 0.12$} & \dpos{$0.26 \pm 0.17$} & \dpos{$0.23 \pm 0.10$} & \dneg{$0.16 \pm 0.09$} \\
    \midrule
    \multicolumn{9}{l}{\emph{(b) Cityscapes}} \\
    Control & $64.07 \pm 1.16$ & $62.23 \pm 0.88$ & $55.90 \pm 0.54$ & $57.89 \pm 0.40$ & $55.16 \pm 0.25$ & $59.85 \pm 0.78$ & $60.95 \pm 0.91$ & $59.44 \pm 0.70$ \\
    \textbf{AnchorUp} & $66.08 \pm 0.29$ & $63.78 \pm 0.17$ & $56.82 \pm 0.04$ & $58.38 \pm 0.09$ & $55.37 \pm 0.19$ & $61.15 \pm 0.12$ & $62.43 \pm 0.22$ & $60.57 \pm 0.08$ \\
    Difference & \dpos{$2.01 \pm 0.87$} & \dpos{$1.55 \pm 0.71$} & \dpos{$0.92 \pm 0.51$} & \dpos{$0.49 \pm 0.47$} & \dpos{$0.21 \pm 0.42$} & \dpos{$1.30 \pm 0.66$} & \dpos{$1.48 \pm 0.69$} & \dpos{$1.14 \pm 0.62$} \\
    \midrule
    \multicolumn{9}{l}{\emph{(c) DAVIS-2017, $\mathcal{J}\&\mathcal{F}$}} \\
    Control & $63.43 \pm 1.45$ & $67.87 \pm 1.15$ & $62.80 \pm 1.14$ & $63.77 \pm 0.90$ & -- & $68.13 \pm 0.59$ & $65.93 \pm 1.10$ & $65.32 \pm 1.05$ \\
    \textbf{AnchorUp} & $68.77 \pm 0.29$ & $72.60 \pm 0.10$ & $65.73 \pm 0.45$ & $65.50 \pm 0.10$ & -- & $71.07 \pm 0.12$ & $69.53 \pm 0.12$ & $68.87 \pm 0.15$ \\
    Difference & \dpos{$5.33 \pm 1.16$} & \dpos{$4.73 \pm 1.24$} & \dpos{$2.93 \pm 0.72$} & \dpos{$1.73 \pm 0.90$} & -- & \dpos{$2.93 \pm 0.47$} & \dpos{$3.60 \pm 1.06$} & \dpos{$3.54 \pm 0.92$} \\
    \bottomrule
  \end{tabular}
  \caption{\textbf{Per-backbone values behind Table~5 of the main paper.} Mean and standard deviation over training seeds 0--2. The control is the third NAF row of Tables~1 and~3 of the main paper ($d = 128$, stage 1 only), and \method{} uses the bilinear anchor; the difference is taken between models of the same seed. DAVIS excludes PEcore, as in Table~3 of the main paper.}
  \label{tab:supp-seeds}
\end{table*}

%% file: tables/tab_ablation.tex
\begin{table*}[t]
  \centering
  \small
  \begin{tabular}{l c c c c c c c c}
    \toprule
    Anchor & $d$ & $\alpha$ & DINOv3 & CLIP & SigLIP~2 & PEcore & Mean & \#Params (M) \\
    \midrule
    Bilinear & 32 & 0.579 & 65.66 & 56.01 & 57.19 & 54.06 & 58.23 & 0.011 \\
     & 64 & 0.607 & 66.09 & 56.53 & 57.81 & 54.69 & 58.78 & 0.043 \\
     & 128 & 0.620 & 66.45 & 57.07 & 58.28 & 55.25 & 59.26 & 0.167 \\
     & 256 & 0.630 & 66.49 & 57.09 & 58.53 & 55.32 & 59.36 & 0.663 \\
    \midrule
    Bicubic & 32 & 0.772 & 65.38 & 56.03 & 57.40 & 54.30 & 58.28 & 0.011 \\
     & 64 & 0.771 & 65.74 & 56.42 & 57.87 & 54.78 & 58.70 & 0.043 \\
     & 128 & 0.771 & 65.53 & 56.78 & 58.58 & 55.67 & 59.14 & 0.167 \\
     & 256 & 0.768 & 66.28 & 56.98 & 58.62 & 55.47 & 59.34 & 0.663 \\
    \bottomrule
  \end{tabular}
  \caption{\textbf{Guidance dimension and anchor.} \method{} trained with the
  recipe of Section~3.5 of the main paper at four guidance dimensions for
  both anchors. Cityscapes mIoU is evaluated by linear probing; the mean is
  calculated over the four backbones shown, not the seven in Table~1 of the
  main paper. The $d = 128$ rows correspond to the models in that table,
  whose costs are reported in Table~4 of the main paper. $\alpha$ is the
  trained gate.}
  \label{tab:ablation}
\end{table*}

%% file: tables/tab_steps.tex
\begin{table}[t]
  \centering
  \small
  \begin{tabular}{l c c c}
    \toprule
    & \multicolumn{2}{c}{Training steps} & \\
    \cmidrule(lr){2-3}
    Benchmark & 25{,}000 & 100{,}000 & Difference \\
    \midrule
    VOC mIoU & 81.06 & 81.08 & $+0.02$ \\
    Cityscapes mIoU & 60.77 & 60.94 & $+0.17$ \\
    DAVIS $\mathcal{J}\&\mathcal{F}$ & 65.90 & 66.29 & $+0.39$ \\
    \bottomrule
  \end{tabular}
  \caption{\textbf{Training length.} \method{} with a bilinear anchor, mean over
  the seven backbones of Table~1 of the main paper. The DAVIS row therefore
  includes PEcore and is not comparable to Table~3 of the main paper, which
  averages over six.
  The main paper reports the 25{,}000-step checkpoint throughout.}
  \label{tab:supp-steps}
\end{table}

%% file: main.bbl
\begin{thebibliography}{29}
\providecommand{\natexlab}[1]{#1}
\providecommand{\url}[1]{\texttt{#1}}
\expandafter\ifx\csname urlstyle\endcsname\relax
  \providecommand{\doi}[1]{doi: #1}\else
  \providecommand{\doi}{doi: \begingroup \urlstyle{rm}\Url}\fi

\bibitem[Bolya et~al.(2025)Bolya, Huang, Sun, Cho, Madotto, Wei, Ma, Zhi,
  Rajasegaran, Bangalath, Wang, Monteiro, Xu, Dong, Ravi, Li, Dollar, and
  Feichtenhofer]{bolya2025pe}
Daniel Bolya, Po-Yao Huang, Peize Sun, Jang~Hyun Cho, Andrea Madotto, Chen Wei,
  Tengyu Ma, Jiale Zhi, Jathushan Rajasegaran, Hanoona Bangalath, Junke Wang,
  Marco Monteiro, Hu Xu, Shiyu Dong, Nikhila Ravi, Shang-Wen Li, Piotr Dollar,
  and Christoph Feichtenhofer.
\newblock {Perception Encoder}: The best visual embeddings are not at the
  output of the network.
\newblock In \emph{Advances in Neural Information Processing Systems
  (NeurIPS)}, pages 67743--67796, 2025.

\bibitem[Chambon et~al.(2026)Chambon, Couairon, Zablocki, Boulch, Thome, and
  Cord]{chambon2026naf}
Lo{\"\i}ck Chambon, Paul Couairon, \'Eloi Zablocki, Alexandre Boulch, Nicolas
  Thome, and Matthieu Cord.
\newblock {NAF}: Zero-shot feature upsampling via neighborhood attention
  filtering.
\newblock In \emph{Conference on Computer Vision and Pattern Recognition
  (CVPR)}, pages 26604--26613, 2026.

\bibitem[Cordts et~al.(2016)Cordts, Omran, Ramos, Rehfeld, Enzweiler, Benenson,
  Franke, Roth, and Schiele]{cordts2016cityscapes}
Marius Cordts, Mohamed Omran, Sebastian Ramos, Timo Rehfeld, Markus Enzweiler,
  Rodrigo Benenson, Uwe Franke, Stefan Roth, and Bernt Schiele.
\newblock The {Cityscapes} dataset for semantic urban scene understanding.
\newblock In \emph{Conference on Computer Vision and Pattern Recognition
  (CVPR)}, pages 3213--3223, 2016.

\bibitem[Couairon et~al.(2025)Couairon, Chambon, Serrano, Haugeard, Cord, and
  Thome]{couairon2025jafar}
Paul Couairon, Lo{\"\i}ck Chambon, Louis Serrano, Jean-Emmanuel Haugeard,
  Matthieu Cord, and Nicolas Thome.
\newblock {JAFAR}: Jack up any feature at any resolution.
\newblock In \emph{Advances in Neural Information Processing Systems
  (NeurIPS)}, pages 110050--110077, 2025.

\bibitem[Deng et~al.(2009)Deng, Dong, Socher, Li, Li, and
  Fei-Fei]{deng2009imagenet}
Jia Deng, Wei Dong, Richard Socher, Li-Jia Li, Kai Li, and Li Fei-Fei.
\newblock {ImageNet}: A large-scale hierarchical image database.
\newblock In \emph{Conference on Computer Vision and Pattern Recognition
  (CVPR)}, pages 248--255, 2009.

\bibitem[Dong et~al.(2014)Dong, Loy, He, and Tang]{dong2014srcnn}
Chao Dong, Chen~Change Loy, Kaiming He, and Xiaoou Tang.
\newblock Learning a deep convolutional network for image super-resolution.
\newblock In \emph{European Conference on Computer Vision (ECCV)}, pages
  184--199, 2014.

\bibitem[Everingham et~al.(2010)Everingham, {Van Gool}, Williams, Winn, and
  Zisserman]{everingham2010voc}
Mark Everingham, Luc {Van Gool}, Christopher K.~I. Williams, John Winn, and
  Andrew Zisserman.
\newblock The {Pascal} visual object classes ({VOC}) challenge.
\newblock \emph{International Journal of Computer Vision}, 88\penalty0
  (2):\penalty0 303--338, 2010.

\bibitem[Fu et~al.(2024)Fu, Hamilton, Brandt, Feldmann, Zhang, and
  Freeman]{fu2024featup}
Stephanie Fu, Mark Hamilton, Laura~E. Brandt, Axel Feldmann, Zhoutong Zhang,
  and William~T. Freeman.
\newblock {FeatUp}: A model-agnostic framework for features at any resolution.
\newblock In \emph{International Conference on Learning Representations
  (ICLR)}, 2024.

\bibitem[He et~al.(2016)He, Zhang, Ren, and Sun]{he2016resnet}
Kaiming He, Xiangyu Zhang, Shaoqing Ren, and Jian Sun.
\newblock Deep residual learning for image recognition.
\newblock In \emph{Conference on Computer Vision and Pattern Recognition
  (CVPR)}, pages 770--778, 2016.

\bibitem[Heinrich et~al.(2025)Heinrich, Ranzinger, Yin, Lu, Kautz, Tao,
  Catanzaro, and Molchanov]{heinrich2024radiov25}
Greg Heinrich, Mike Ranzinger, Hongxu~Danny Yin, Yao Lu, Jan Kautz, Andrew Tao,
  Bryan Catanzaro, and Pavlo Molchanov.
\newblock {RADIOv2.5}: Improved baselines for agglomerative vision foundation
  models.
\newblock In \emph{Conference on Computer Vision and Pattern Recognition
  (CVPR)}, pages 22487--22497, 2025.

\bibitem[Heo et~al.(2024)Heo, Park, Han, and Yun]{heo2024rope}
Byeongho Heo, Song Park, Dongyoon Han, and Sangdoo Yun.
\newblock Rotary position embedding for vision transformer.
\newblock In \emph{European Conference on Computer Vision (ECCV)}, pages
  289--305, 2024.

\bibitem[Huang et~al.(2025)Huang, Chen, Havrylov, Geiger, and
  Zhang]{huang2025loftup}
Haiwen Huang, Anpei Chen, Volodymyr Havrylov, Andreas Geiger, and Dan Zhang.
\newblock {LoftUp}: Learning a coordinate-based feature upsampler for vision
  foundation models.
\newblock In \emph{International Conference on Computer Vision (ICCV)}, pages
  9913--9923, 2025.

\bibitem[Kim et~al.(2016)Kim, Lee, and Lee]{kim2016vdsr}
Jiwon Kim, Jung~Kwon Lee, and Kyoung~Mu Lee.
\newblock Accurate image super-resolution using very deep convolutional
  networks.
\newblock In \emph{Conference on Computer Vision and Pattern Recognition
  (CVPR)}, pages 1646--1654, 2016.

\bibitem[Kopf et~al.(2007)Kopf, Cohen, Lischinski, and
  Uyttendaele]{kopf2007jbu}
Johannes Kopf, Michael~F. Cohen, Dani Lischinski, and Matt Uyttendaele.
\newblock Joint bilateral upsampling.
\newblock \emph{ACM Transactions on Graphics}, 26\penalty0 (3):\penalty0 96,
  2007.

\bibitem[Liu et~al.(2023)Liu, Lu, Fu, and Cao]{liu2023dysample}
Wenze Liu, Hao Lu, Hongtao Fu, and Zhiguo Cao.
\newblock Learning to upsample by learning to sample.
\newblock In \emph{International Conference on Computer Vision (ICCV)}, pages
  6004--6014, 2023.

\bibitem[Loshchilov and Hutter(2019)]{loshchilov2019adamw}
Ilya Loshchilov and Frank Hutter.
\newblock Decoupled weight decay regularization.
\newblock In \emph{International Conference on Learning Representations
  (ICLR)}, 2019.

\bibitem[Lu et~al.(2022{\natexlab{a}})Lu, Liu, Fu, and Cao]{lu2022fade}
Hao Lu, Wenze Liu, Hongtao Fu, and Zhiguo Cao.
\newblock {FADE}: Fusing the assets of decoder and encoder for task-agnostic
  upsampling.
\newblock In \emph{European Conference on Computer Vision (ECCV)}, pages
  231--247, 2022{\natexlab{a}}.

\bibitem[Lu et~al.(2022{\natexlab{b}})Lu, Liu, Ye, Fu, Liu, and
  Cao]{lu2022sapa}
Hao Lu, Wenze Liu, Zixuan Ye, Hongtao Fu, Yuliang Liu, and Zhiguo Cao.
\newblock {SAPA}: Similarity-aware point affiliation for feature upsampling.
\newblock In \emph{Advances in Neural Information Processing Systems
  (NeurIPS)}, pages 20889--20901, 2022{\natexlab{b}}.

\bibitem[Pont-Tuset et~al.(2017)Pont-Tuset, Perazzi, Caelles, Arbel\'aez,
  Sorkine-Hornung, and {Van Gool}]{pont2017davis}
Jordi Pont-Tuset, Federico Perazzi, Sergi Caelles, Pablo Arbel\'aez, Alexander
  Sorkine-Hornung, and Luc {Van Gool}.
\newblock The 2017 {DAVIS} challenge on video object segmentation.
\newblock \emph{arXiv preprint arXiv:1704.00675}, 2017.

\bibitem[Radford et~al.(2021)Radford, Kim, Hallacy, Ramesh, Goh, Agarwal,
  Sastry, Askell, Mishkin, Clark, Krueger, and Sutskever]{radford2021clip}
Alec Radford, Jong~Wook Kim, Chris Hallacy, Aditya Ramesh, Gabriel Goh,
  Sandhini Agarwal, Girish Sastry, Amanda Askell, Pamela Mishkin, Jack Clark,
  Gretchen Krueger, and Ilya Sutskever.
\newblock Learning transferable visual models from natural language
  supervision.
\newblock In \emph{International Conference on Machine Learning (ICML)}, pages
  8748--8763, 2021.

\bibitem[Ramachandran et~al.(2019)Ramachandran, Parmar, Vaswani, Bello,
  Levskaya, and Shlens]{ramachandran2019sasa}
Prajit Ramachandran, Niki Parmar, Ashish Vaswani, Irwan Bello, Anselm Levskaya,
  and Jonathon Shlens.
\newblock Stand-alone self-attention in vision models.
\newblock In \emph{Advances in Neural Information Processing Systems
  (NeurIPS)}, pages 68--80, 2019.

\bibitem[Ranzinger et~al.(2024)Ranzinger, Heinrich, Kautz, and
  Molchanov]{ranzinger2024amradio}
Mike Ranzinger, Greg Heinrich, Jan Kautz, and Pavlo Molchanov.
\newblock {AM-RADIO}: Agglomerative vision foundation model reduce all domains
  into one.
\newblock In \emph{Conference on Computer Vision and Pattern Recognition
  (CVPR)}, pages 12490--12500, 2024.

\bibitem[Sim\'eoni et~al.(2026)Sim\'eoni, Vo, Seitzer, Baldassarre, Oquab,
  Jose, Khalidov, Szafraniec, Yi, Ramamonjisoa, Massa, Haziza, Wehrstedt, Wang,
  Darcet, Moutakanni, Sentana, Roberts, Vedaldi, Tolan, Brandt, Couprie,
  Mairal, J\'egou, Labatut, and Bojanowski]{simeoni2025dinov3}
Oriane Sim\'eoni, Huy~V. Vo, Maximilian Seitzer, Federico Baldassarre, Maxime
  Oquab, Cijo Jose, Vasil Khalidov, Marc Szafraniec, Seungeun Yi, Micha\"el
  Ramamonjisoa, Francisco Massa, Daniel Haziza, Luca Wehrstedt, Jianyuan Wang,
  Timoth\'ee Darcet, Th\'eo Moutakanni, Leonel Sentana, Claire Roberts, Andrea
  Vedaldi, Jamie Tolan, John Brandt, Camille Couprie, Julien Mairal, Herv\'e
  J\'egou, Patrick Labatut, and Piotr Bojanowski.
\newblock {DINOv3}.
\newblock \emph{Transactions on Machine Learning Research}, 2026.

\bibitem[Su et~al.(2024)Su, Ahmed, Lu, Pan, Bo, and Liu]{su2021roformer}
Jianlin Su, Murtadha Ahmed, Yu Lu, Shengfeng Pan, Wen Bo, and Yunfeng Liu.
\newblock {RoFormer}: Enhanced transformer with rotary position embedding.
\newblock \emph{Neurocomputing}, 568:\penalty0 127063, 2024.

\bibitem[Tschannen et~al.(2025)Tschannen, Gritsenko, Wang, Naeem,
  Alabdulmohsin, Parthasarathy, Evans, Beyer, Xia, Mustafa, H\'enaff, Harmsen,
  Steiner, and Zhai]{tschannen2025siglip2}
Michael Tschannen, Alexey Gritsenko, Xiao Wang, Muhammad~Ferjad Naeem, Ibrahim
  Alabdulmohsin, Nikhil Parthasarathy, Talfan Evans, Lucas Beyer, Ye Xia, Basil
  Mustafa, Olivier H\'enaff, Jeremiah Harmsen, Andreas Steiner, and Xiaohua
  Zhai.
\newblock {SigLIP} 2: Multilingual vision-language encoders with improved
  semantic understanding, localization, and dense features.
\newblock \emph{arXiv preprint arXiv:2502.14786}, 2025.

\bibitem[Venkataramanan et~al.(2026)Venkataramanan, Pariza, Salehi, Knobel,
  Ramzi, Gidaris, Bursuc, and Asano]{venkataramanan2025franca}
Shashanka Venkataramanan, Valentinos Pariza, Mohammadreza Salehi, Lukas Knobel,
  Elias Ramzi, Spyros Gidaris, Andrei Bursuc, and Yuki~M. Asano.
\newblock Franca: Nested matryoshka clustering for scalable visual
  representation learning.
\newblock In \emph{Conference on Computer Vision and Pattern Recognition
  (CVPR)}, pages 10533--10544, 2026.

\bibitem[Walmer et~al.(2026)Walmer, Suri, Aggarwal, and
  Shrivastava]{walmer2026uplift}
Matthew Walmer, Saksham Suri, Anirud Aggarwal, and Abhinav Shrivastava.
\newblock {UPLiFT}: Efficient pixel-dense feature upsampling with local
  attenders.
\newblock In \emph{Conference on Computer Vision and Pattern Recognition
  (CVPR)}, pages 41288--41298, 2026.

\bibitem[Wang et~al.(2019)Wang, Chen, Xu, Liu, Loy, and Lin]{wang2019carafe}
Jiaqi Wang, Kai Chen, Rui Xu, Ziwei Liu, Chen~Change Loy, and Dahua Lin.
\newblock {CARAFE}: {Content-Aware} {ReAssembly} of {FEatures}.
\newblock In \emph{International Conference on Computer Vision (ICCV)}, pages
  3007--3016, 2019.

\bibitem[Wimmer et~al.(2026)Wimmer, Truong, Rakotosaona, Oechsle, Tombari,
  Schiele, and Lenssen]{wimmer2025anyup}
Thomas Wimmer, Prune Truong, Marie-Julie Rakotosaona, Michael Oechsle, Federico
  Tombari, Bernt Schiele, and Jan~Eric Lenssen.
\newblock {AnyUp}: Universal feature upsampling.
\newblock In \emph{International Conference on Learning Representations
  (ICLR)}, 2026.

\end{thebibliography}
